\documentclass[10pt,twocolumn,letterpaper]{article}

\usepackage{cvpr}              %

\usepackage{graphicx}
\usepackage{amsmath}
\usepackage{amssymb}
\usepackage{booktabs}

\usepackage[dvipsnames]{xcolor}
\usepackage{amsfonts}
\usepackage{bm}
\usepackage{amsmath}
\usepackage{mathtools}
\usepackage{multirow}
\usepackage{booktabs}
\usepackage{caption}
\usepackage{enumitem}

\renewcommand{\paragraph}[1]{{\vspace{1mm}\noindent \bf #1}.}

\newcommand{\X}{\mathbf{X}}

\newcommand{\Y}{\mathbf{Y}}
\newcommand{\Ygt}{Y_\text{gt}}
\newcommand{\y}{\mathbf{y}}
\newcommand{\yt}{\y_t}
\newcommand{\map}{\mathcal{M}}
\newcommand{\Ego}{\mathbf{Y}^\text{plan}}
\newcommand{\ego}{\mathbf{y}^\text{plan}}

\newcommand{\Egoapprox}{\hat{\mathbf{Y}}^\text{plan}}
\newcommand{\egoapprox}{\hat{\mathbf{y}}^\text{plan}}
\newcommand{\egoapproxt}{\egoapprox_t}
\newcommand{\z}{\mathbf{z}}
\newcommand{\zplan}{\mathbf{\z^\text{plan}}}
\newcommand{\h}{\mathbf{h}}

\newcommand{\Lcvae}{\mathcal{L}_\text{cvae}}
\newcommand{\Ladv}{\mathcal{L}_\text{adv}}
\newcommand{\Lrecon}{\mathcal{L}_\text{recon}}
\newcommand{\LKL}{\mathcal{L}_\text{KL}}
\newcommand{\Lcoll}{\mathcal{L}_\text{coll}}
\newcommand{\Lenv}{\mathcal{L}_\text{env}}
\newcommand{\Lveh}{\mathcal{L}_\text{agent}}
\newcommand{\Lprior}{\mathcal{L}_\text{prior}}
\newcommand{\Linit}{\mathcal{L}_\text{init}}
\newcommand{\Laccel}{\mathcal{L}_\text{accel}}

\newcommand{\reals}{\mathbb{R}}

\newcommand{\normal}{\mathcal{N}}

\newcommand{\w}[1]{w_\text{#1}}

\newcommand{\A}{\mathbf{A}}
\newcommand{\accel}{\mathbf{a}}

\newcommand{\name}{{{STRIVE}}\xspace}

\usepackage[pagebackref,breaklinks,colorlinks]{hyperref}

\usepackage[capitalize]{cleveref}
\crefname{section}{Sec.}{Secs.}
\Crefname{section}{Section}{Sections}
\Crefname{table}{Table}{Tables}
\crefname{table}{Tab.}{Tabs.}

\def\cvprPaperID{****} %
\def\confName{CVPR}
\def\confYear{2022}

\begin{document}
\setlength{\abovedisplayskip}{5pt}
\setlength{\belowdisplayskip}{5pt}

\title{Generating Useful Accident-Prone Driving Scenarios via a Learned Traffic Prior}

\author{Davis Rempe$^{1,2}$\qquad Jonah Philion$^{2,3,4}$\qquad Leonidas J. Guibas$^{1}$ \qquad Sanja Fidler$^{2,3,4}$ \qquad Or Litany$^{2}$\\ \vspace{1mm}
\text{\normalsize $^1$Stanford University\qquad $^2$NVIDIA \qquad $^3$University of Toronto \qquad $^4$Vector Institute }\\
\href{https://nv-tlabs.github.io/STRIVE}{\texttt{nv-tlabs.github.io/STRIVE}}
}

\maketitle

\begin{abstract}
Evaluating and improving planning for autonomous vehicles requires scalable generation of long-tail traffic scenarios. To be useful, these scenarios must be realistic and challenging, but not impossible to drive through safely. In this work, we introduce STRIVE, a method to automatically generate challenging scenarios that cause a given planner to produce undesirable behavior, like collisions. To maintain scenario plausibility, the key idea is to leverage a learned model of traffic motion in the form of a graph-based conditional VAE. Scenario generation is formulated as an optimization in the latent space of this traffic model, perturbing an initial real-world scene to produce trajectories that collide with a given planner. A subsequent optimization is used to find a ``solution" to the scenario, ensuring it is useful to improve the given planner. Further analysis clusters generated scenarios based on collision type. We attack two planners and show that STRIVE successfully generates realistic, challenging scenarios in both cases. We additionally ``close the loop'' and use these scenarios to optimize hyperparameters of a rule-based planner.
\end{abstract}
\vspace{-6mm}
\section{Introduction}
\label{sec:intro}

\begin{figure}[t]
\vspace{-3mm}
\begin{center}
\includegraphics[width=1.0\linewidth]{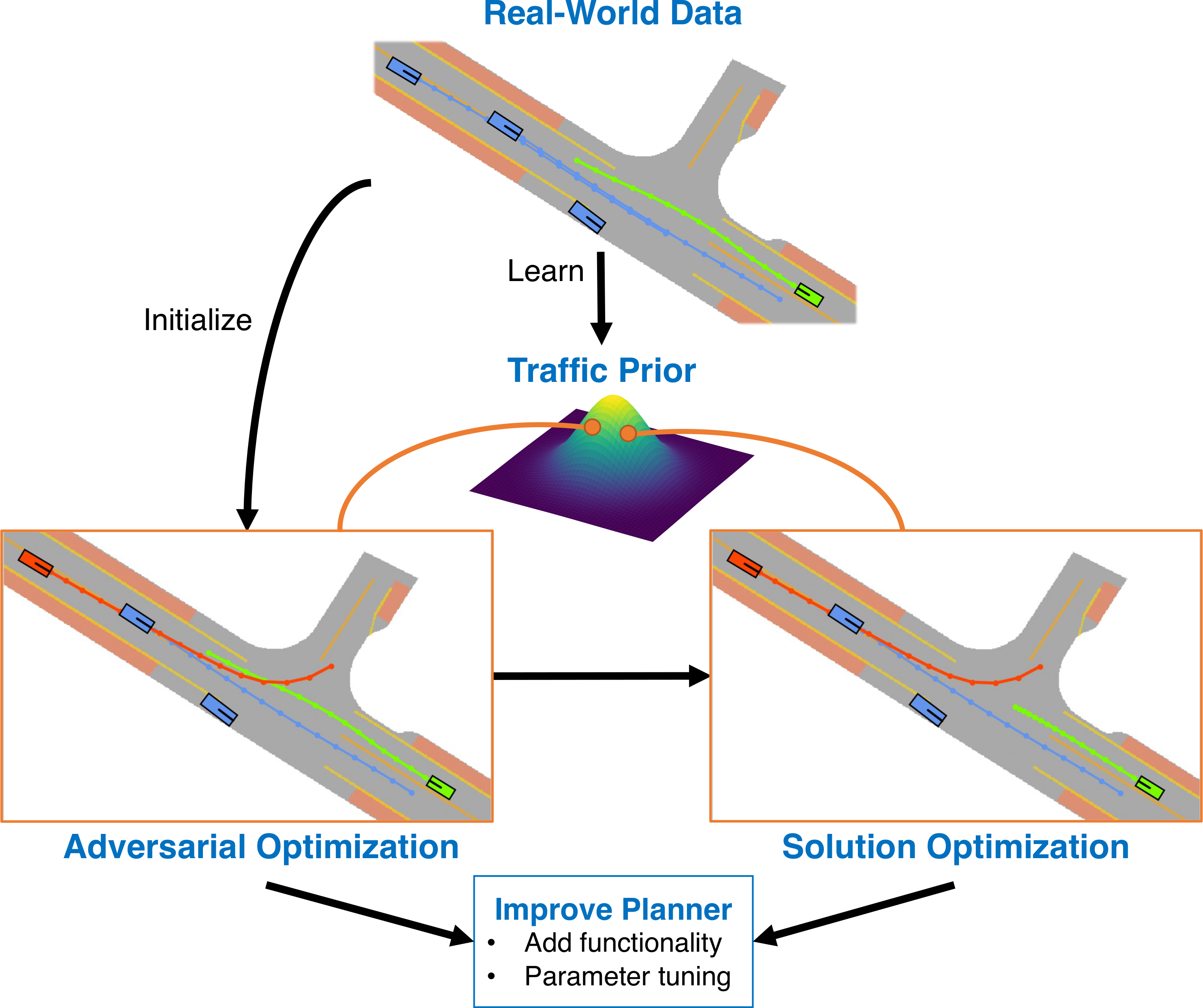}
\end{center}
\vspace{-5.5mm}
   \caption{\footnotesize \name generates challenging scenarios for a given planner. An \emph{adversarial optimization} perturbs a real-world scene in the latent space of a learned traffic model, causing an adversary (red) to collide with the planner (green). A subsequent \emph{solution optimization} finds a planner trajectory to avoid collisions, verifying a scenario is useful for identifying planner improvements.
   }
\vspace{-4.5mm}
\label{fig:overview}
\end{figure}

The safety of contemporary autonomous vehicles (AVs) is defined by their ability to safely handle complicated near-collision scenarios.
However, these kinds of scenarios are rare in real-world driving, posing a data-scarcity problem
that is detrimental to both the development and testing of  data-driven models for perception, prediction, and planning.
Moreover, the better models become, the more rare these events will be, making the models even harder to train.

A natural solution is to synthesize difficult scenarios in simulation, rather than relying on real-world data, making it easier and safer to evaluate and train AV systems. This approach is especially appealing for \textit{planning}, where the appearance domain gap is not a concern. 
For example, one can manually design scenarios where the AV may fail by inserting adversarial actors or modifying trajectories, either from scratch or by perturbing a small set of real scenarios.
Unfortunately, the manual nature of this approach quickly becomes prohibitively expensive when a large set of scenarios is necessary for training or comprehensive evaluation.

Recent work looks to \textit{automatically generate} challenging scenarios~\cite{wang2021advsim,abeysirigoonawardena2019generating,ding2021multimodal,ding2020learning,vemprala2020adversarial,o2018scalable,klischat2019generating}. 
Generally, these approaches control a single or small group of ``adversaries'' in a scene, define an objective (\eg cause a collision with the AV), and then optimize the adversaries' behavior or trajectories to meet the objective.
While most methods demonstrate generation of only 1 or 2 scenarios~\cite{abeysirigoonawardena2019generating,chen2021adversarial,klischat2019generating,o2018scalable}, recent work~\cite{wang2021advsim} has improved scalability by starting from real-world traffic scenes and perturbing a limited set of pre-chosen adversaries.
However, these approaches \emph{lack expressive priors over plausible traffic motion}, which limits the realism and diversity of scenarios.
In particular, adversarial entities in a scenario are a small set of agents heuristically chosen ahead of time; surrounding traffic will not be reactive and therefore perturbations must be careful to avoid implausible situations (\eg collisions with auxiliary agents).
Furthermore, less attention has been given to determining if a scenario is ``unsolvable''~\cite{ghodsi2021generating}, \ie, if even an oracle AV is incapable of avoiding a collision. 
In this degenerate case, the scenario is not \emph{useful} for evaluating/training a planner.

In this work, we introduce \name~-- a method for generating challenging scenarios to \textbf{S}tress-\textbf{T}est d\textbf{RIVE} a given AV system. 
\name attacks the prediction, planning, and control subset of the AV stack, which we collectively refer to as the \emph{planner}.
As shown in \cref{fig:overview}, our approach perturbs an initial real-world scene through an optimization procedure to cause a collision between an arbitrary adversary and a given planner.
Our core idea is to measure the plausibility of a scenario during optimization by its likelihood under a learned generative model of traffic motion, %
which encourages scenarios to be challenging, yet realistic.
As a result, \name does not choose specific adversaries ahead of time, rather it jointly optimizes all scene agents, enabling a diverse set of scenarios to arise. 
Moreover, in order to accommodate for non-differentiable (or inaccessible) planners, which are widely used in practice, the proposed optimization uses a differentiable proxy representation of the planner within the learned motion model, thus allowing standard gradient-based optimization to be used.

We propose to identify and characterize generated scenarios that are \emph{useful} for improving a given planner. We first search for a ``solution'' to generated scenarios to determine if they are degenerate, and then cluster solvable scenarios based on collision properties. %
We test \name on two AV planners, including a new rule-based planner, and show that it generates plausible and diverse collision scenarios in both cases.
We additionally use generated scenarios to improve the rule-based planner by identifying fundamental limitations of its design and tuning hyperparameters.

In short, our contributions are: (i) a method to automatically generate plausible challenging scenarios for a given planner, (ii) a solution optimization to ensure scenario utility, and (iii) an analysis method to cluster scenarios by collision type.
Supplementary videos and material for this work are available on the \href{https://nv-tlabs.github.io/STRIVE}{project webpage}.

\vspace{-1mm}
\section{Related Work}
\label{sec:relwork}

\vspace{-1mm}
\paragraph{Traffic Motion Modeling} Scenario replay is insufficient for testing and developing AV planners as the motion of non-ego vehicles is strongly coupled to the actions chosen by the ego planner. Advances in deep learning have allowed us to replace traditional dynamic and kinematic models~\cite{wan2000unscented,kong2015bicycle,lefevre2014survey} or rule-based simulators~\cite{lopez2018microscopic,dosovitskiy2017carla} with neural counterparts that better capture traffic complexity \cite{bansal2018chauffeurnet,pal2020emergent}.  
Efforts to predict future trajectories from a short state history and an HD map can generally be categorized according to the encoding technique, modeling of multi-modality, multi-agent interaction, and whether the trajectory is estimated in a single step or progressively. 
The encoding of surrounding context of each agent is often done via a bird's-eye view (BEV) raster image~\cite{cui2019multimodal,chai2019multipath,fang2020tpnet}, though some work~\cite{liang2020learning,gao2020vectornet} replaces the rasterization-based map encoding with a lane-graph representation. SimNet~\cite{bergamini2021simnet} increased the diversity of generated simulations by initializing the state using a generative model conditioned on the semantic map.
To account for multi-modality, multiple futures have been estimated either directly \cite{cui2019multimodal} or through trajectory proposals \cite{chai2019multipath,fang2020tpnet, liu2021multimodal,learneval20}. %
Modeling multi-actor interactions explicitly using dense graphs has proven effective for vehicles~\cite{casas2020implicit, suo2021trafficsim}, lanes~\cite{liang2020learning}, and pedestrians~\cite{ivanovic2019trajectron, salzmann2020trajectron++, kosaraju2019social}. Finally, step-by-step prediction has performed favorably to one-shot prediction of the whole trajectory~\cite{djuric2020uncertainty}. We follow these works and design a traffic model that uses an inter-agent graph network \cite{johnson2020learning} to represent agent interaction and is variational, allowing us to sample multiple futures. 

Our model builds on VAE-based approaches~\cite{casas2020implicit,suo2021trafficsim} that provide a learned prior over a controllable latent space~\cite{rempe2021humor}. Among other design differences, we incorporate a penalty for environment collisions and structure predictions through a bicycle model to ensure physical plausibility. 

\paragraph{Challenging Scenario Generation}
Generating scenarios has the potential to exponentially increase scene coverage compared to relying exclusively on recorded drives. 
Advances in photo-realistic simulators like CARLA~\cite{dosovitskiy2017carla} and NVIDIA's DRIVE Sim, along with the availability of large-scale datasets~\cite{caesar2020nuscenes,sun2020scalability,ettinger2021large,kesten2020lyftperception,houston2020lyftprediction}, have been instrumental to methods that generate plausible scene graphs to improve perception~\cite{kar2019meta,devaranjan2020metasim2,Resnick_2021_ICCV} and planning~\cite{bergamini2021simnet,casas2020implicit,suo2021trafficsim,Kim2021_DriveGAN}.
Our work focuses on generating challenging -- or ``adversarial''\footnote{we use ``challenging'' to denote generation procedures that do not explicitly attack a specific module in the perception or planning stack} -- scenarios, which are even more crucial since they are so rare in recorded data. While most works assume perfect perception and attack the planning module~\cite{chen2021adversarial,klischat2019generating,ding2021multimodal,ding2020learning,vemprala2020adversarial,ghodsi2021generating}, recent efforts exploit the full stack, including image or point-cloud perception~\cite{abeysirigoonawardena2019generating,o2018scalable,li2021fooling,wang2021advsim,tu2020physically}.
Our work focuses on attacking the planner only, though our scene parameterization as a learned traffic model could be incorporated into end-to-end methods. Unlike our approach, which uses gradient-based optimization enabled by the learned motion model, most adversarial generation works rely heavily on black-box optimization which may be slow and unreliable.

Our scenario generation approach is most similar to AdvSim~\cite{wang2021advsim}, however instead of optimizing acceleration profiles of a simplistic bicycle model we use a more expressive data-driven motion prior. This remedies the previous difficulty of controlling many adversarial agents simultaneously in a plausible manner. Moreover, we avoid constraining the attack trajectory to not collide with the playback AV by  proposing a ``solution'' optimization stage to filter worthwhile scenarios. 
Prior work \cite{chen2021adversarial} clusters lane-change scenarios based on trajectories of agents, while we cluster based on collision properties between the adversary and planner.

\paragraph{AV Planners}
Despite the recent academic interest in end-to-end learning-based planners and AVs~\cite{sadat2019jointly,zeng2019end,sadat2020perceive,casas2021mp3,bansal2018chauffeurnet}, rule-based planners remain the norm in practical AV systems~\cite{vitelli2021safetynet}. 
Therefore, we evaluate our approach on a rule-based planner similar to the lane-graph-based planners used by successful teams in the 2007 DARPA Urban Challenge~\cite{stanforddarpa,Urmson-2007-9708} detailed in \cref{sec:methodplanner}.

\vspace{-2mm}
\section{Challenging Scenario Generation}
\label{sec:method}
\name aims to generate high-risk traffic situations for a given \emph{planner}, which can subsequently be used to improve that planner (\cref{fig:overview}).
For our purpose, the planner encapsulates prediction, planning, and control, \ie we're interested in scenarios where the system misbehaves even with perfect perception.
The planner takes as input past trajectories of other agents in a scene and outputs the future trajectory of the vehicle it controls (termed the \emph{ego} vehicle).
It is assumed to be black-box: \name has no knowledge of the planner's internals and cannot compute gradients through it.
Undesirable behavior includes collisions with other vehicles and non-drivable terrain, uncomfortable driving (\eg high accelerations), and breaking traffic laws.
We focus on generating \textbf{accident-prone scenarios} involving vehicle-vehicle collisions with the planner, though our formulation is general and in principle can handle alternative objectives.

\begin{figure*}
\begin{center}
\includegraphics[width=1.0\textwidth]{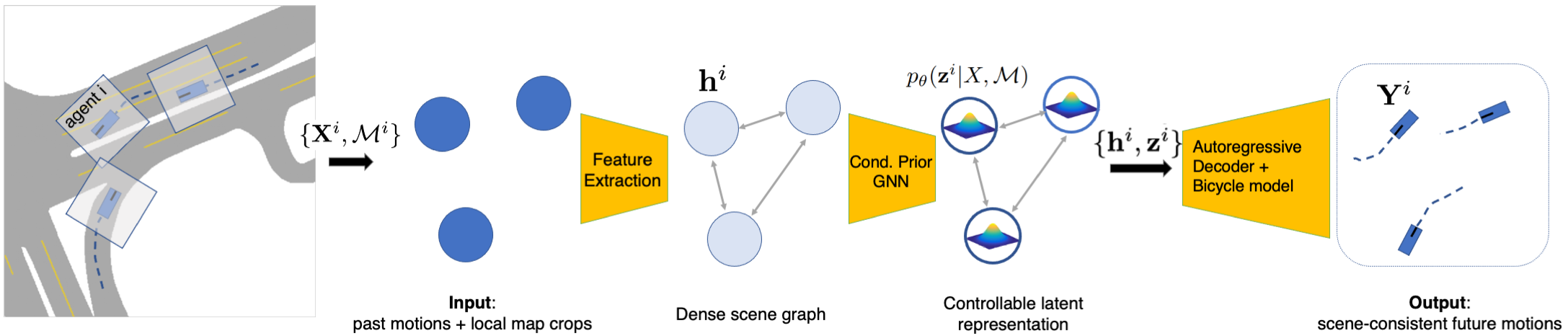}
\end{center}
\vspace{-5mm}
   \caption{Test-time architecture of the learned traffic model. To jointly sample future trajectories for all agents in a scene, past motion and local map context is first processed individually for each agent. The \emph{conditional prior}, then outputs a latent distribution at each node that can be sampled and fed through the \emph{autoregressive decoder} to predict future agent trajectories.
   }
\label{fig:arch}\vspace{-3mm}
\end{figure*}

Similar to prior work~\cite{wang2021advsim}, scenario generation is formulated as an optimization problem that perturbs agent trajectories in an initial scenario from real-world data.
The input is a planner $f$, map $\map$ containing semantic layers for drivable area and lanes, and a sequence from a pre-recorded real-world scene that serves as initialization for optimization. %
This initial scenario contains $N$ agents with trajectories represented in 2D BEV as $Y = \{\Y^i\}_{i=1}^{N}$, where $\Y^i = [ \y_1^i, \y_2^i, \dots, \y_{T}^i]$ is the sequence of states for agent $i$. 
We let $\Y_t = [ \y_t^1, \y_t^2, \dots, \y_t^N]$ be the state of all agents at a single timestep.
An agent state $\yt^i = [x_t, y_t, \theta_t, v_t, \Dot{\theta}_t]$ at time $t$ contains the 2D position $(x_t,y_t)$, heading $\theta_t$, speed $v_t$, and yaw rate $\Dot{\theta}_t$. 
When rolled out within a scenario, at each timestep the planner outputs the next \textbf{ego} state $\ego_t = f (\ego_{<t}, \Y_{<t}, \map)$ based on the \emph{past} motion of itself and other agents. 
For simplicity, we will write the rolled out planner trajectory as $\Ego = f(Y, \map)$ where $\Ego = [ \ego_1, \ego_2, \dots, \ego_{T}]$ for the remainder of this paper. %
Scenario generation perturbs trajectories for all non-ego agents to best meet an adversarial objective $\Ladv$ (\eg cause a collision with the planner):
\begin{equation}
    \min_{Y} \Ladv(Y, \Ego), \quad \Ego = f(Y, \map).
\end{equation}

One may optimize a single or small set of ``adversaries'' in $Y$ explicitly, \eg through the kinematic bicycle model parameterization~\cite{wang2021advsim,kong2015bicycle,polack2017bicycle}. 
While this enforces plausible single-agent dynamics, \textbf{interactions} must be constrained to avoid collisions between non-ego agents and, even then, the resulting traffic patterns may be unrealistic. 
We propose to instead \emph{learn} to model traffic motion using a neural network and then use it at optimization time \textbf{(i) to parameterize} all trajectories in a scenario as vectors in the latent space, and \textbf{(ii) as a prior} over scenario plausibility.
Next, we describe this traffic model, followed by the ``adversarial'' optimization that produces collision scenarios.

\subsection{Modeling ``Realism'': Learned Traffic Model}
We wish to generate accident-prone scenarios that are assumed to develop over short time periods (\textless 10 sec)~\cite{najm2007pre}. 
Therefore, traffic modeling is formulated as \emph{future forecasting}, which predicts future trajectories for all agents in a scene based on their past motion.
We learn $p_\theta(Y | X, \map)$ to enable sampling a future scenario $Y$ conditioned on the fixed past $X = \{\X^i\}_{i=1}^{N}$ (defined similar to $Y$) and the map $\map$.
Two properties of the traffic model make it particularly amenable to downstream optimization: a low-dimensional latent space for efficient optimization, and a prior distribution over this latent space to determine the plausibility of a given scenario.
Inspired by recent work~\cite{casas2020implicit,suo2021trafficsim}, we design a conditional variational autoencoder (CVAE), shown in \cref{fig:arch}, that meets these criteria while learning accurate and scene-consistent joint future predictions. 
We briefly introduce the architecture and training procedure here, and refer to the supplement for specific details.

\paragraph{Architecture}
To sample future motions at test time, the \emph{conditional prior} and \emph{decoder} are used; both are graph neural networks (GNN) operating on a fully-connected scene graph of all agents.
The \emph{prior} models $p_\theta(Z | X, \map)$ where $Z = \{\z^i\}_{i=1}^{N}$ is a set of agent latent vectors.
Each node in the input scene graph contains a context feature $\h^i$ extracted from that agent's past trajectory, local rasterized map, bounding-box size, and semantic class.
After message passing, the \emph{prior} outputs parameters of a Gaussian $ p_\theta(\z^i | X, \map) = \normal (\mu_\theta^i(X, \map), \sigma_\theta^i(X, \map))$ for each agent in the scene, forming a ``distributed'' latent representation that captures the variation in possible futures.

The deterministic \emph{decoder} $Y = d_\theta(Z, X, \map)$ operates on the scene graph with both a sampled latent $\z^i$ and past context $\h^i$ at each node.
Decoding is performed autoregressively: at timestep $t$, one round of message passing resolves interactions before predicting accelerations $\Dot{v}_t, \Ddot{\theta}_t$ for each agent.
Accelerations immediately go through the kinematic bicycle model~\cite{kong2015bicycle,polack2017bicycle} to obtain the next state  $\y_{t+1}^i$, which updates $\h^i$ before continuing rollout.
The determinism and graph structure of the decoder encourages scene-consistent future predictions even when agent $\z$'s are independently sampled.
Importantly for latent optimization, the decoder ensures plausible vehicle dynamics by using the kinematic bicycle model, even if the input $Z$ is unlikely.

\paragraph{Training}
Training is performed on pairs of $(X, \Ygt)$ using a modified CVAE objective: %
\begin{equation}
    \Lcvae = \Lrecon + \w{KL} \LKL + \w{coll}\Lcoll.
\label{cvaeloss}
\end{equation}
To optimize this loss, a \emph{posterior} network $q_\phi(Z | \Ygt, X, \map)$ is introduced similar to the prior, but operating jointly on past and future motion.
Future trajectory features are extracted separately, while past features are the same as used in the \emph{prior}.
The full training loss uses trajectory samples from both the posterior $Y_\text{post}$ and prior $Y_\text{prior}$:\\[-2mm]
\begin{align}
    \Lrecon &= \sum_{i=1}^{N} || \Y_\text{post}^i - \Y_\text{gt}^i ||^2 \\
    \LKL &= D_\text{KL} (q_\phi(Z | \Ygt, X, \map) || p_\theta(Z | X, \map)) \\
    \Lcoll &= \Lveh + \Lenv
\end{align}
where $\Y_\text{post}^i \in Y_\text{post}$, $\Y_\text{gt}^i \in Y_\text{gt}$, and $D_\text{KL}$ is the KL divergence.
Collision penalties $\Lveh$ and $\Lenv$ use a differentiable approximation of collision detection as in \cite{suo2021trafficsim}, which represents vehicles by sets of discs to penalize $Y_\text{prior}$ for collisions between agents or with the non-drivable map area.

\subsection{Adversarial Optimization}
\label{sec:advopt}
To leverage the learned traffic model, the real-world scenario used to initialize optimization is split into the past $X$ and future $Y_\text{init}$.
Throughout optimization, past trajectories in $X$ (including that of the planner) are \emph{fixed} while the future is perturbed to cause a collision with the given planner $f$.
This perturbation is done in the learned latent space of the traffic model -- as described below, we optimize the set of latents for all $N$ \textit{non-ego} agents $Z = \{\z^i\}_{i=1}^{N}$ along with a latent representation of the planner $\zplan$.

Latent scenario parameterization encourages plausibility in two ways.
First, since the decoder is trained on real-world data, it will output realistic traffic patterns if $Z$ stays near the learned manifold.
Second, the learned prior network gives a distribution over latents, which is used to penalize unlikely $Z$. 
This strong prior on behavioral plausibility enables jointly optimizing \emph{all agents} in the scene rather than choosing a small set of specific adversaries in advance.

At each step of optimization (\cref{fig:optimarch}), the perturbed scenario is decoded with $d_\theta(Z, \zplan, X, \map)$ and non-ego trajectories $Y$ are passed to the (black-box) planner, which rolls out the ego motion before losses can be computed.
Adversarial optimization seeks two simultaneous objectives:

\begin{figure}[t]
  \centering
  \includegraphics[width=\linewidth]{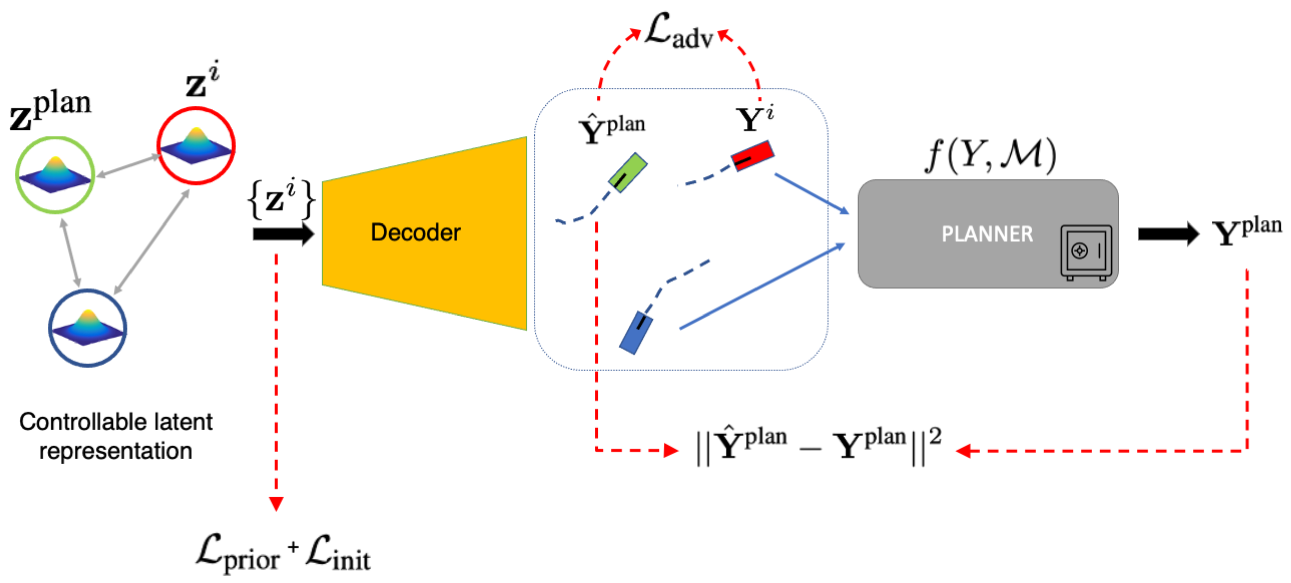}
  \vspace{-4mm}
  \caption{At each step of adversarial optimization, latent representations of both the planner and non-ego agents are decoded with the learned decoder and non-ego trajectories are given to the planner for rollout within the scenario. Finally, losses are computed.}
  \vspace{-5mm}
  \label{fig:optimarch}
\end{figure}

\paragraph{1. Match Planner}\label{par:matchplanner}
Although optimization has no direct control over the planner's behavior -- an external function that is queried only when required -- it is still necessary to represent the planner within the traffic model (\ie include it in the scene graph with an associated latent $\zplan$) so that interactions with other agents are realistic.
In doing this, future predictions from the decoder include an estimate of the planner trajectory $\Egoapprox$ that, ideally, is close to the true planner trajectory $\Ego = f(Y, \map)$. 
Note that this gives a \emph{differentiable approximation} of the planner, enabling typical gradient-based optimization to be used for the second objective described below.
To encourage matching the real planner output with this ``internal'' approximation, we use
\begin{equation}
    \min_\zplan || \Egoapprox - \Ego ||^2 - \alpha \log p_\theta(\z^\text{plan} | X, \map)
\label{eqn:tgtmatch}
\end{equation}
where the right term lightly regularizes $\zplan$ to stay likely under the learned prior and $\alpha$ balances the two terms.

\paragraph{2. Collide with Planner}
The goal for non-ego agents is to cause the planner to collide with another vehicle:
\begin{equation}
    \min_{Z} \Ladv + \Lprior + \Linit + \Lcoll.
\end{equation}

\noindent The \textit{adversarial} term encourages a collision by minimizing the positional distance between controlled agents and the current traffic model approximation of the planner: 
\begin{align}
    \Ladv &= \sum_{i=1}^N \sum_{t=1}^T \delta_t^i \cdot || \yt^i - \egoapproxt ||^2 \label{eqn:lossadv}
 \\
    \delta_t^i &= \dfrac{\exp(-|| \yt^i - \egoapproxt ||)}{\sum_j \sum_t \exp(-|| \yt^j - \egoapproxt ||)}
\label{eqn:softmin}
\end{align}
where $\yt$ here only includes the 2D position.
Intuitively, the $\delta_t^i$ coefficients defined by the \textit{softmin} in \cref{eqn:softmin} are finding a candidate agent and timestep to collide with the planner.
The agent with the largest $\delta_t^i$ is the most likely ``adversary'' based on distance, and \cref{eqn:lossadv} prioritizes causing a collision between this adversary and the planner while still allowing gradients to reach other agents.
This weighting helps $\Lprior$ to avoid all agents unrealistically colliding with the planner.

The \textit{prior} term encourages latents to stay likely under the learned prior network:
\begin{align}
    \Lprior &= - \dfrac{1}{N} \sum_{i=1}^N \gamma^i \cdot \log p_\theta(\z^i | X, \map) \\
    \gamma^i &= 1 - \sum_t \delta_t^i.
\label{eqn:priorweight}
\end{align}
The $\gamma^i$ coefficient will weight likely adversaries near zero, \ie agents close to colliding with the planner are allowed to deviate from the learned traffic manifold to exhibit rare and challenging behavior.
Because the traffic model training data does not contain collisions, we found it difficult for an agent to collide with the planner using a large prior loss, thus motivating the weighting in \cref{eqn:priorweight}.
Note that even when $\gamma^i$ is small, agents will maintain physical plausibility since the decoder uses the kinematic bicycle model.

$\Linit$ encourages staying close to the initialization in latent space, since it is already known to be realistic:
\begin{align}
    \Linit = \dfrac{1}{N} \sum_{i=1}^N \gamma^i \cdot || \z^i - \z_\text{init}^i ||^2
\end{align}
where $\z_\text{init}^i \in Z_\text{init}$ are the latents that initialize optimization.
\noindent Finally, similar to CVAE training, $\Lcoll$ discourages non-ego agents from colliding with each other and the non-drivable area. 
In practice, all loss terms are balanced by manual inspection of a small set of generated scenarios.

\paragraph{Initialization and Optimization}
Given a real-world scene, $Z_\text{init}$ is obtained through the posterior network $q_\phi$, then further refined with an initialization optimization that fits to the input future trajectories of all agents (similar to \cref{eqn:tgtmatch}), including the initial planner rollout.
Optimization is implemented in PyTorch\cite{paszke2017automatic} using ADAM\cite{kingma2014adam} with a learning rate of $0.05$. 
Runtime depends on the planner and number of agents; for our rule-based planner (see \cref{sec:methodplanner}), a 10-agent scenario takes 6-7 minutes.

\section{Analyzing and Using Generated Scenarios}

\subsection{Filtering and Collision Classification}
\label{sec:solopt}
\paragraph{Solution Optimization}
Adversarial optimization produces plausible scenarios, but it cannot guarantee they are \textit{solvable} and \textit{useful}: \eg a scenario in which the ego is squeezed by multiple cars produces an unavoidable collision and is therefore uninformative for evaluating or improving a planner. %
Therefore, we perform an additional optimization to identify an ego trajectory that avoids collision; if this optimization fails, the scenario is discarded for downstream tasks.
This solution optimization is initialized from the output of the adversarial optimization and essentially inverts the objectives described in \cref{sec:advopt}: non-ego latent $Z$ are tuned to maintain the adversarial trajectories while $\z^\text{plan}$ is optimized to avoid collisions and stay likely under the prior. 

\paragraph{Clustering and Labeling}
To gain insight into the distribution of collision scenarios and inform their downstream use, we propose a simple approach to cluster and label them.
Specifically, scenarios are characterized by the explicit relationship between the planner and adversary at the time of collision: the relative direction and heading of the adversary are computed in the frame of the planner and concatenated to form a collision feature for each scenario. %
These features are clustered with $k$-means to form semantically similar groups of accidents that are labeled by visual inspection. %
Their distribution can then be visualized as in \cref{fig:analysis}.

\vspace{-2mm}
\subsection{Improving the Planner}
\label{sec:methodplanner}
With a large set of labeled collision scenarios, the planner can be improved in two main ways.
First, discrete improvements to functionality may be needed if many scenarios of the same type are generated.
For example, a planner that strictly follows lanes is subject to collisions from head on or behind as it fails to swerve, indicating necessary new functionality to leave the lane graph.
Second, scenarios provide data for tuning hyperparameters or learned parameters.

\paragraph{Rule-based Planner}
To demonstrate how \name scenarios are used for these kinds of improvements, we introduce a simple, yet competent, rule-based planner that we use as a proxy for a real-world planner. 
Our planner is ideal for evaluating STRIVE as it is easily interpretable, uses a small set of hyperparameters, and has known failure modes.
In short, it relies entirely on the lane graph to both predict future trajectories of non-ego vehicles and generate candidate trajectories for the ego vehicle.
Among these candidates, it chooses that which covers the most distance with a low ``probability of collision.''
Planner behavior is affected by hyperparameters such as maximum speed/acceleration and how collision probability is computed.
This planner has the additional limitation that it cannot change lanes, which scenarios generated by \name exposes in \cref{sec:improveplanner}.
Full details of this planner are included in the supplementary. %

\vspace{-3mm}
\section{Experiments}
\label{sec:results}
\vspace{-1mm}
We next highlight the new capabilities that \name enables.
\cref{sec:resultsadvgen} demonstrates the ability to generate challenging and useful scenarios on two different planners; these scenarios contain a diverse set of collisions, as shown through analysis in \cref{sec:resultsanalysis}.
Generated scenarios are used to improve our rule-based planner in \cref{sec:improveplanner}.

\paragraph{Dataset}
The nuScenes dataset~\cite{caesar2020nuscenes} is used both to train the traffic model and to initialize adversarial optimization.
It contains $20s$ traffic clips annotated at 2 Hz, which we split into $8s$ scenarios.
Only \textit{car} and \textit{truck} vehicles are used and the traffic model operates on the rasterized \textit{drivable area}, \textit{carpark area}, \textit{road divider}, and \textit{lane divider} map layers.
We use the splits and settings of the nuScenes prediction challenge which is $2s$ (4 steps) of past motion to predict $6s$ (12 steps) of future, meaning collision scenarios are $8s$ long, but only the future $6s$ trajectories are optimized.

\paragraph{Planners}
Scenario generation is evaluated on two different planners. 
The \emph{Replay} planner simply plays back the ground truth ego trajectory from nuScenes data.
This is an open-loop setting where the planner's $6s$ future is fully rolled out without re-planning.
The \emph{Rule-based} planner, described in \cref{sec:methodplanner}, allows a more realistic closed-loop setting where the planner reacts to the surrounding agents during future rollout by re-planning at 5 Hz.

\paragraph{Metrics} 
The \textbf{collision rate} is the fraction of optimized initial scenarios from nuScenes that succeed in causing a planner collision, which indicates the sample efficiency of scenario generation. \textbf{Solution rate} is the fraction of these colliding scenarios for which a solution was found, which measures how often scenarios are \emph{useful}.
\textbf{Acceleration} indicates how comfortable a driven trajectory is; challenging scenarios should generally increase acceleration for the planner, while the adversary's acceleration should be reasonably low to maintain plausibility.
If a scenario contains a collision, acceleration (and other trajectory metrics) is only calculated up to the time of collision.
\textbf{Collision velocity} is the relative speed between the planner and adversary at the time of collision; it points to the severity of a collision.

\begin{figure}[t]
\begin{center}
\includegraphics[width=1.0\linewidth]{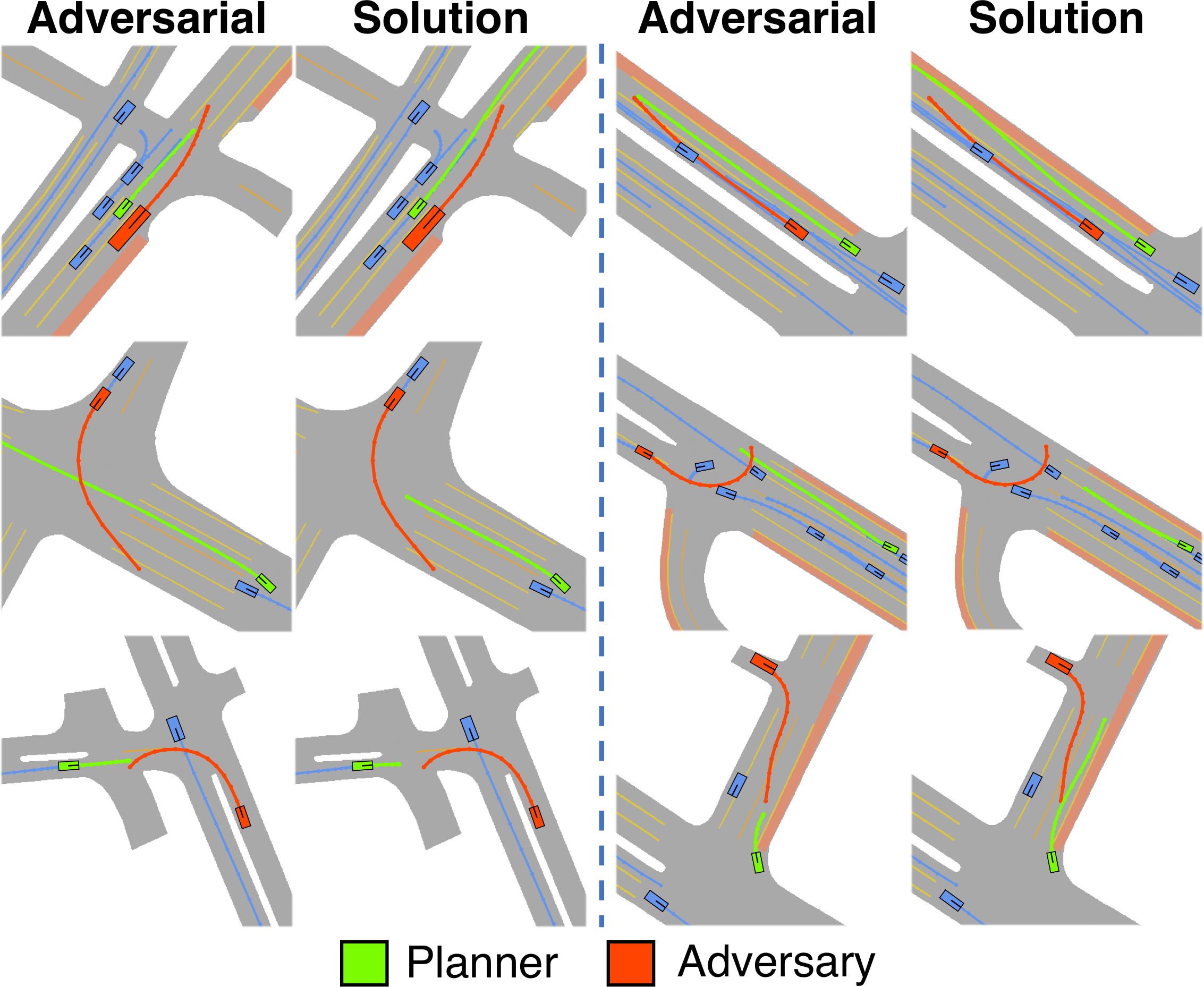}
\end{center}
\vspace{-5mm}
   \caption{Qualitative results on the \emph{Rule-based} planner. Adversarial and solution optimization results are shown. \name produces diverse collision scenarios including lane changes, (u-)turning in front of the planner, and pulling into oncoming traffic.
   }
\vspace{-5mm}
\label{fig:qualitative}
\end{figure}
\begin{table*}
\begin{center}
\scalebox{0.95}{
\begin{tabular}{ll|lc|ll|cc}
\toprule
 & \multicolumn{1}{c}{} & \multicolumn{2}{c}{\textbf{}} & \multicolumn{2}{c}{\textbf{Planner Trajectory}} & \multicolumn{2}{c}{\textbf{Match Planner Err}} \\
\textbf{Planner} & \multicolumn{1}{c}{\textbf{Scenarios}} & \multicolumn{1}{c}{Collision \small{(\%)}} & \multicolumn{1}{c}{Solution \small{(\%)}} & Accel \small{($m/s^2$)} & \multicolumn{1}{c}{Coll Vel \small{($m/s$)}} & Pos \small{(m)} & Ang \small{(deg)} \\
\midrule
Replay & Challenging & 43.7 \small{(\textbf{+43.7})} & 82.4 & 0.85 & 7.82 & 0.28 & 1.32 \\
\midrule
Rule-based & Regular & 1.2 & -- & 1.63 & 8.48 & -- & -- \\
Rule-based & Challenging & 27.4 \small{(\textbf{+26.2})} & 86.8 & 1.91 \small{(\textbf{+0.28})} & 9.65 \small{(\textbf{+1.17})} & 1.23 & 3.79 \\
\bottomrule
\end{tabular}}
\end{center}
\vspace{-5mm}
\caption{Evaluation of generated \textit{challenging} scenarios. Generated scenarios contain far more collisions compared to the corresponding \textit{regular} (unmodified) scenes, as well as higher acceleration and collision speeds. Acceleration is measured in the forward direction (\ie change in speed), since the \emph{Rule-based} planner cannot change lanes. Rightmost columns show small errors between $\Egoapprox$ and $\Ego$.
}
\label{tab:advgen}
\vspace{-4mm}
\end{table*}
\begin{table}
\setlength{\tabcolsep}{3pt}
\begin{center}
\scalebox{0.8}{
\begin{tabular}{l|cccc|c}
\toprule
 \multicolumn{1}{c}{} & \multicolumn{4}{c}{\textbf{Plausibility of Adversary Trajectory} $\downarrow$} & \multicolumn{1}{c}{\textbf{Usefulness}}$\uparrow$ \\
\multicolumn{1}{c}{\textbf{Scenarios}} & Accel \small{($m/s^2$)} & \multicolumn{1}{c}{Env Coll \small{(\%)}} & NN Dist \small{(m)} & \multicolumn{1}{c}{NLL} & \multicolumn{1}{c}{Solution \small{(\%)}} \\
\midrule
Bicycle &  2.00 & 16.5 & 0.97 & 962.9 & 73.4 \\
\name & \textbf{0.98} & \textbf{10.8} & \textbf{0.72} & \textbf{323.4} & \textbf{83.5} \\
\bottomrule
\end{tabular}}
\end{center}
\vspace{-5mm}
\caption{Scenario generation for \emph{Replay} planner compared to the \emph{Bicycle} baseline, which does not leverage a learned traffic model.}
\label{tab:baselines}
\vspace{-5mm}
\end{table}        

\vspace{-1mm}
\subsection{Scenario Generation Evaluation}
\label{sec:resultsadvgen}
\vspace{-1mm}
First, we demonstrate that \name generates challenging, yet solvable, scenarios causing planners to collide and drive uncomfortably.
Moreover, compared to an alternative generation approach that \emph{does not} leverage the learned traffic prior, \name scenarios are more plausible.
Scenario generation is initialized from 1200 $8s$ sequences from nuScenes.
Before adversarial optimization, scenes are pre-filtered heuristically 
by how likely they are to produce a useful collision, leaving \textless500 scenarios to optimize.

\paragraph{Planner-Specific Scenarios}
\cref{tab:advgen} shows that compared to rolling out a given planner on ``regular'' (unmodified) nuScenes scenarios, challenging scenarios from \name produce more collisions and less comfortable driving.
For the \emph{Rule-based} planner, metrics on challenging scenarios are compared to the corresponding set of regular scenarios from which they originated (regular scenarios for \emph{Replay} are omitted since nuScenes data contains no collisions and, by definition, planner behavior does not change).

Collision and solution rates indicate that generated scenarios are accident-prone and \textit{useful} (solvable).
For the \emph{Rule-based} planner, adversarial optimization causes collisions in $27.4\%$ of scenarios compared to only $1.2\%$ in the regular scenarios. %
Generated scenarios also contain more severe collisions in terms of velocity, and elicit larger accelerations, \ie less comfortable driving.
The position and angle errors between approximate ($\Egoapprox$) and true ($\Ego$) planner trajectories at the end of adversarial optimization are shown on the right (see~\cref{par:matchplanner}).
The largest position error of $1.23m$ is reasonable relative to the $4.084m$ length of the planner vehicle. %
Qualitative results for the \emph{Rule-based} planner visualize 2D waypoint trajectories (\cref{fig:qualitative}); though not shown, \name also generates speed and heading.

\paragraph{Baseline Comparison}
\name is next compared to a baseline approach to demonstrate that 
leveraging a learned traffic model is key to realistic and useful scenarios.
Previous works are not directly comparable as they focus on small-scale scenario generation (\eg \cite{abeysirigoonawardena2019generating,chen2021adversarial}) and/or attack the full AV stack rather than just the planner~\cite{wang2021advsim,o2018scalable}.
Therefore, in the spirit of AdvSim~\cite{wang2021advsim} we implement the \emph{Bicycle} baseline, which explicitly optimizes the kinematic bicycle model parameters (acceleration profile) of a single pre-chosen adversary in the scenario to cause a collision.
Rather than using the learned traffic model, it relies on the bicycle model, collision penalties, and acceleration regularization to maintain plausibility.
This precludes using the differentiable planner approximation from the traffic model, thus requiring gradient estimation (\eg finite differences) for the closed-loop setting, which we found is $\approx40\times$ slower and requires several hours to generate a single scenario.
Therefore, comparison is done only on the \emph{Replay} planner.

\begin{figure}[t]
\begin{center}
\includegraphics[width=1.0\linewidth]{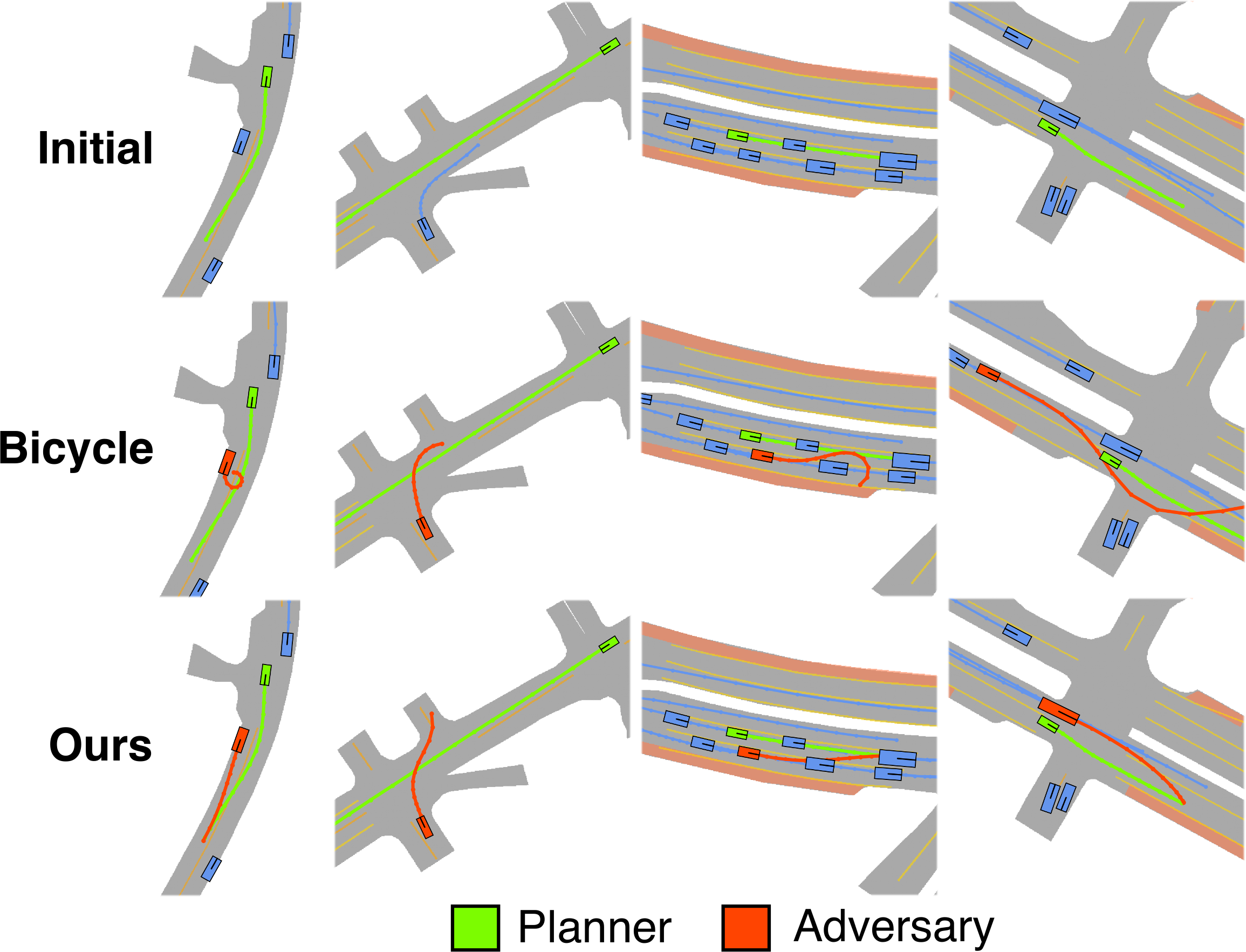}
\end{center}
\vspace{-5mm}
   \caption{Qualitative comparison of generated scenarios for the \emph{Replay} planner. \emph{Bicycle} often produces semantically unrealistic trajectories as no learned traffic model is leveraged.
   }
\vspace{-5mm}
\label{fig:baselines}
\end{figure}

\cref{tab:baselines} shows that scenarios generated by \emph{Bicycle} exhibit more unrealistic adversarial driving, and are more difficult to find a solution for.
All metrics are reported only for scenarios where both methods successfully caused a collision.
In addition to higher accelerations, the \emph{Bicycle} adversary collides with the non-drivable area more often (\emph{Env Coll}), and exhibits less typical trajectories as measured by the distance to the nearest-neighbor ego trajectory in the nuScenes training split (\emph{NN Dist}).
After fitting the \emph{Bicycle}-generated scenarios with our traffic model, we see the adversary's behavior is also less realistic as measured by the negative log-likelihood (NLL) of its latent $\z$ under the learned prior.
These observations are supported qualitatively in \cref{fig:baselines}.

\subsection{Analyzing Generated Scenarios}
\label{sec:resultsanalysis}
\begin{figure*}
\begin{center}
\includegraphics[width=0.95\textwidth]{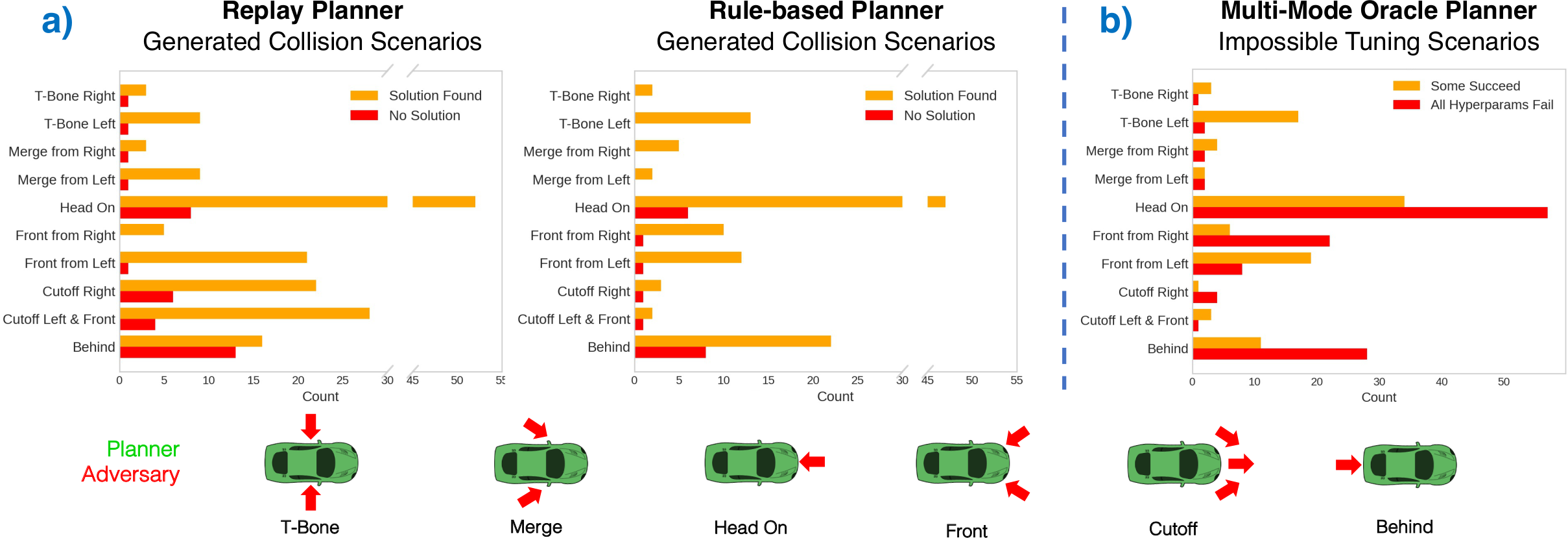}
\end{center}
\vspace{-5mm}
   \caption{(Bottom) Collision types are depicted; the arrow indicates the position/direction of the adversary. \textbf{(a)} Distribution of generated scenarios for both planners. \textbf{(b)} Scenarios used to tune the multi-mode \emph{oracle} planner. Scenarios where all parameter settings cause a collision are in red. A majority of \textit{Head On}, \textit{Front from Right}, and \textit{Behind} scenarios always fail due to the inability to change lanes.
   }
\vspace{-3mm}
\label{fig:analysis}
\end{figure*}

\vspace{-1mm}
Before improving a given planner, the analysis from \cref{sec:solopt} is used to identify useful scenarios by filtering out unsolvable scenarios and classifying collision types.
For classification, collision features are clustered with $k=10$ and clusters are visualized to manually assign the semantic labels shown in \cref{fig:analysis}.
The distribution of generated collision scenarios for both planners in \cref{sec:resultsadvgen} is shown in \cref{fig:analysis}(a) (see supplement for visualized examples from clusters).
\name generates a diverse set of scenarios with solvable scenes found in all clusters.
``Head On'' is the most frequently generated scenario type, likely because \emph{Replay} is non-reactive and \emph{Rule-based} cannot change lanes.
``Behind'' exhibits the highest rate of unsolvable scenarios since being hit from behind is often the result of a negligent following vehicle, rather than undesirable planner behavior.
\emph{Replay} is much more susceptible to being cut off since it is open-loop, while the closed-loop \emph{Rule-based} can successfully react to avoid such collisions.

\subsection{Improving Rule-Based Planner}
\label{sec:improveplanner}
\vspace{-1mm}
Now that we have a large set of labeled collision scenarios, in addition to the original nuScenes data containing ``regular'' scenarios (with few collisions), we can improve the \emph{Rule-based} planner to be better prepared for challenging situations.
Besides uncovering fundamental flaws that lead us to add new functionality, improvement is based on hyperparameter tuning via a grid search over possible settings.
For each set of hyperparameters, the planner is rolled out within all scenarios of a dataset, and the optimal tuning is chosen based on the minimum collision rate.

The planner is first tuned on regular scenarios before adversarial optimization is performed to create a set of challenging scenarios to guide further improvements.
Performance of this initial regular-tuned planner on held out regular (\emph{Reg}) and collision (\emph{Coll}) scenarios is shown in the top row of \cref{tab:tuneplanner}.
Before any improvements, the planner collides in $68.6\%$ of challenging and $4.6\%$ of regular scenarios.
Note that avoiding collisions altogether on regular scenarios is not possible: even if we choose optimal hyperparameters for each scenario separately, the collision rate is still $3.2\%$.

\begin{table}
\setlength{\tabcolsep}{3pt}
\begin{center}
\scalebox{0.85}{
\begin{tabular}{l|ccc}
\toprule
\multicolumn{1}{c}{} & \multicolumn{1}{c}{\textbf{Collision} (\%)} & \textbf{Coll Vel} ($m/s$) & \textbf{Accel} ($m/s^2$)  \\
\multicolumn{1}{c}{\textbf{Improvement}} & Reg / Coll & Reg / Coll & Reg / Coll \\
\midrule
None (regular-tuned) & 4.6 / 68.6 & 4.59 / 10.48 & 1.96 / 2.26 \\
+ Challenging data & 6.0 / 51.4  & 5.48 / 13.88 & 2.29 / 2.50  \\
+ Extra \emph{learned} mode & 4.6 / 54.3 &  4.60 / 10.86 & 2.02 / 2.55 \\
\midrule
+ Extra \emph{oracle} mode & 4.6 / 54.3 & 4.59 / 10.40 & 1.96 / 2.39 \\
\bottomrule
\end{tabular}}
\end{center}
\vspace{-5mm}
\caption{Improving \emph{Rule-Based} planner. Including challenging tuning data and adding an extra ``mode'' improves performance on collision scenarios (\emph{Coll}) while maintaining performance in regular scenarios (\emph{Reg}). Acceleration is in the forward direction.}
\vspace{-5mm}
\label{tab:tuneplanner}
\end{table}

\paragraph{Tuning on Challenging Scenes}
The first improvement, shown in the second row of \cref{tab:tuneplanner}, is to na{\"i}vely combine regular and challenging scenarios for tuning.
Combined tuning greatly reduces the collision rate on challenging scenarios, but negatively impacts performance on regular driving.
This points to a first \emph{fundamental issue}: the planner uses a single set of hyperparameters for all driving situations, %
causing it to drive too aggressively in regular scenarios when tuned on challenging ones.

\paragraph{Multi-Mode Operation}
To address this, we add a second set of parameters such that the planner has one for regular and one for accident-prone situations.
Using this second ``accident mode'' of operation requires a binary classification of the current scene during rollout. 
For this, we augment the planner with a learned component that decides which parameter set to use based on a moving window of the past $2s$ of traffic; it is trained on scenarios generated by \name.
The extra parameter set is tuned on collision scenarios only.
As shown in the third row of \cref{tab:tuneplanner}, this learned extra mode reduces the collision rate on challenging scenarios by $14.3\%$ compared to the vanilla planner without hindering performance on regular scenes.
We compare it to an \emph{oracle} version (bottom row) that automatically switches into accident mode $2s$ before a collision is supposed to happen on generated scenarios, showing the learned version is achieving near-optimal performance.

\paragraph{Lane Change Limitation}
The inability of the planner to switch lanes is another fundamental issue exposed by collision scenarios.
\cref{fig:analysis}(b) shows the distribution of tuning scenarios for the \emph{oracle} multi-mode version; red bars indicate ``impossible'' scenarios where all sets of evaluated hyperparameters collide.
A majority of ``Head On'' and ``Behind'' scenarios are impossible, pointing out the lane change limitation.
Adversarial optimization has indeed exploited the flaw and the proposed analysis made it visible.

\vspace{-1mm}
\section{Discussion}
\vspace{-1mm}
\name enables automatic and scalable generation of plausible, accident-prone scenarios to improve a given planner.
However, remaining limitations offer potential future directions.
Our method assumes perfect perception and only attacks the planner, but using our traffic model to additionally attack detection and tracking is of great interest.
\name generates scenarios from existing data and only considers collisions between vehicles, but other incidents involving pedestrians and cyclists are also important, and other kinds of adversaries like adding/removing assets and changing map topology will uncover additional AV weaknesses.
Our method is intended to make AVs safer by exposing them to challenging and rare scenarios similar to the real world.
However, our experiments expose the difficulty of properly balancing regular and challenging data when tuning a planner.
Care must be taken to integrate generated scenarios into AV testing and to design unified planners that robustly address highly variable driving conditions.

{\small \paragraph{Acknowledgments} 
Author Guibas was supported by a Vannevar Bush faculty fellowship.
The authors thank Amlan Kar for the fruitful discussion and feedback throughout the project.
}

{\small
\bibliographystyle{ieee_fullname}
\bibliography{egbib}
}

\clearpage
\appendix
\section*{Appendices}
Here we include additional technical details and results omitted from the main paper for brevity.
\cref{sec:supp-methoddetails} gives details on the main technical methods, \cref{sec:supp-exptdetails} gives details for experiments, and \cref{sec:supp-results} provides additional results to supplement those in the main paper.

\paragraph{Video Results} We encourage viewing the video results on the \href{https://nv-tlabs.github.io/STRIVE}{project page} to get a better sense of the kinds of scenarios produced by \name. 

\section{Method Details} \label{sec:supp-methoddetails}

In this section, we provide additional technical and implementation details about the methods introduced in Sec 3, 4, and 5 of the main paper.
Note that due to many formulated optimization problems being very similar, mathematical notation is often overloaded and may mean slightly different things depending on the section/optimization problem. This is always noted in the text.

\subsection{Learned Traffic Model}
\label{sec:supptrafficarch}
The learned traffic model is introduced in Sec 3.1 of the main paper.
Here we step through the main components of the architecture in more detail.
All neural network components use ReLU activations and layer normalization~\cite{ba2016layer} unless noted otherwise.

\paragraph{State Representation}
In practice, trajectories used as input and supervision for the traffic model are represented as $\Y^i = [ \y_1^i, \y_2^i, \dots, \y_{T}^i]$ for agent $i$ with $T$ timesteps where the state at time $t$ is $\yt^i = [x_t, y_t, \theta_t^x, \theta_t^y, v_t, \Dot{\theta}_t] \in \reals^6$ containing the 2D position $(x_t,y_t)$, heading unit vector $(\theta_t^x, \theta_t^y)$, speed $v_t$, and yaw rate $\Dot{\theta}_t$.

\paragraph{Feature Extraction}
Context features for each agent in the scene are first extracted based on:  the past trajectory $\X^i$, the map $\map$, 
a one-hot encoding of the semantic class $\mathbf{s}^i$ (either \emph{car} or \emph{truck} for the experiments in the main paper), and the agent's bounding box length/width $\mathbf{b}^i = (l, w)$.
The past trajectory feature for each agent $\mathbf{p}^i \in \reals^{64}$ is encoded from $\X^i$, $\mathbf{s}^i$, and $\mathbf{b}^i$ using a 4-layer MLP with hidden size 128. To handle missing frames where an agent has not been annotated in nuScenes~\cite{caesar2020nuscenes}, the past encoding MLP is also given a flag for each input timestep indicating whether the agent state at that step is valid, \ie whether it is true nuScenes data or has been filled with dummy zeros.
The map feature $\mathbf{m}^i \in 64$ is extracted from a local map crop of size $256\times256$ around the agent ($17m$ behind, $60m$ in front, $38.5m$ to each side) at the last step of $\X^i$.
The map input contains one channel for each layer in the input (\textit{drivable area}, \textit{carpark area}, \textit{road divider}, and \textit{lane divider} in nuScenes); each layer is a binary image rasterized at 4 pixels/$m$.
The map extraction network is a CNN with 6 convolutional layers (stride 2, kernel sizes $[7, 5, 5, 3, 3, 3]$, and number of filters $[16, 32, 64, 64, 128, 128]$) followed by a single fully-connected layer.
Map extraction uses group normalization between layers~\cite{wu2018group}.

\paragraph{Prior Network}
The input to the prior is a fully-connected scene graph with a context feature placed at each node that is the concatenation of the previously detailed features $\h^i = [\mathbf{p}^i$, $\mathbf{m}^i, \mathbf{s}^i]$.
Before message passing, each $\h^i$ is further processed with a small 3-layer input MLP to be size 128.

The prior, posterior, and decoder are all graph neural networks (GNN) similar to a scene interaction module~\cite{casas2020implicit,suo2021trafficsim}.
They perform one round of message passing, which involves an \emph{edge network}, \emph{aggregation function}, and \emph{update network}.
Consider a single node $i$ in the scene graph.
First, interaction features are computed for every incoming edge.
For an edge from node $j \rightarrow i$, the edge feature is computed using the \emph{edge network} $\mathcal{E}$ as $\mathbf{e}^{ij} = \mathcal{E}(\h^i, \h^j, \mathcal{T}^{ij})$ where $\mathcal{T}^{ij}$ is the relative position and heading of agent $j$ in the local frame of agent $i$.
$\mathcal{E}$ is a 3-layer MLP with hidden and output size of 128.
After computing all edge features, they are aggregated into a single interaction feature $\mathbf{e}^i \in \reals^{128}$ using maxpooling $\mathbf{e}^i = \max (\mathbf{e}^{i1}, \mathbf{e}^{i2}, \dots)$.
The update network then gives the output at each node $\mathbf{o}^i = \mathcal{U}(\h^i, \mathbf{e}^i)$; it is a 4-layer MLP with hidden size 128.

In the case of the prior network, the outputs at each node are the parameters of a Gaussian distribution.
In particular, $\mathbf{o}^i = [\mu^i, \sigma^i]$ where $\mu^i \in \reals^{32}$ gives the mean and $\sigma^i \in \reals^{32}$ parameterizes the diagonal covariance matrix so that $ p_\theta(\z^i | X, \map) = \normal (\mu^i_\theta(X, \map), \sigma^i_\theta(X, \map))$ where $\theta$ are the learned parameters of the prior network.

\begin{figure}[t]
  \centering
  \includegraphics[width=\linewidth]{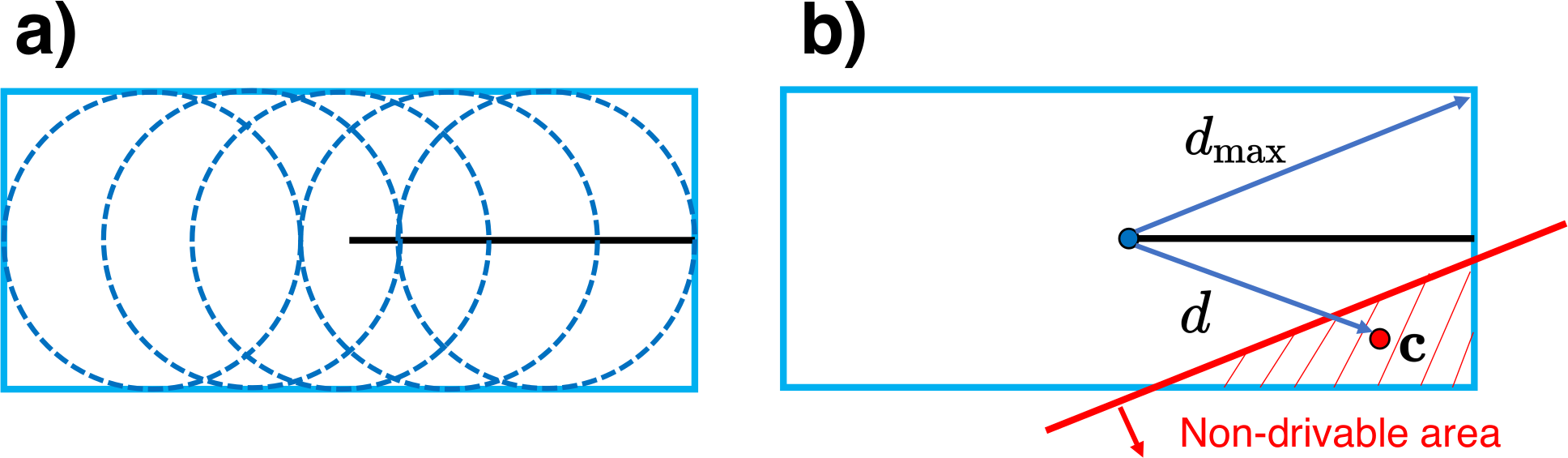}
  \vspace{-4mm}
  \caption{Collision penalties. (a) Computing $\Lveh$ estimates each vehicle by a set of 5 discs. (b) $\Lenv$ is computed based on the distance between a collision point $\mathbf{c}$ and the vehicle position.}
  \vspace{-5mm}
  \label{fig:collloss}
\end{figure}

\paragraph{Posterior (Encoder) Network}
The approximate posterior network (commonly referred to as the \emph{encoder} in CVAEs) is used (i) during training and (ii) to initialize adversarial optimization.
It is nearly the same as the prior, but takes in an additional future feature extracted from the future trajectory for each agent.
The future trajectory feature for each agent $\mathbf{f}^i \in \reals^{64}$ is encoded from $\Y_\text{gt}^i$, $\mathbf{s}^i$, and $\mathbf{b}^i$ using a 4-layer MLP with hidden size 128.
Node $i$ of the scene graph input to the posterior contains a concatenation of $[\mathbf{f}^i, \mathbf{p}^i$, $\mathbf{m}^i, \mathbf{s}^i]$ where $\mathbf{p}^i$, $\mathbf{m}^i$ are the same past and map features given to the prior network.
Input processing and message passing is performed in the same way as the prior, and the final output is also a distribution over latents $ q_\phi(\z^i | Y_\text{gt}, X, \map) = \normal (\mu^i_\phi(Y_\text{gt}, X, \map), \sigma^i_\phi(Y_\text{gt}, X, \map))$ with $\phi$ the learned parameters of the network.

\paragraph{Decoder Network}
As described in the main paper, the decoder progresses autoregressively, predicting one future step at a time.
When predicting step $t$, each node of the input scene graph contains a concatenation of $[\z^i, \mathbf{p}^i_{t-1}$, $\mathbf{m}^i_{t-1}, \mathbf{s}^i, \mathbf{b}^i]$ where $\z^i$ is sampled from either the prior or posterior output.
The past and map features are updated throughout autoregressive rollout, and therefore notated by a time subscript; initially these are the exact same as given to the prior, \ie $\mathbf{p}^i_0 = \mathbf{p}^i$ and $\mathbf{m}^i_0 = \mathbf{m}^i$.

At step $t$ of rollout, the concatenated input features are first processed by a 3-layer input MLP (hidden size 128) to get a single 64-dimensional feature at each node.
A single round of message passing proceeds in the same way as for the prior to get the output $\mathbf{o}^i_t = [\Dot{v}_t^i, \Ddot{\theta}_t^i] \in \reals^2$ at each node, which contains current linear and angular acceleration.
The kinematic bicycle model~\cite{kong2015bicycle,polack2017bicycle} is then used to get the actual agent state $\yt^i = \mathcal{K}(\y_{t-1}^i, \mathbf{o}^i_t, \mathbf{b}^i)$.
Before proceeding to the next step of rollout, the past and map context features must be updated according to the new state.
In particular, the past feature is updated using a gated-recurrent unit RNN $\mathbf{p}^i_{t} = \text{GRU}(\mathbf{p}^i_{t-1}, \yt^i)$ with 3 layers (the GRU uses a hidden state of size 64 that is omitted here for brevity). The new map feature $\mathbf{m}^i_{t}$ is extracted using the same CNN as before, but with an updated map crop in the local frame of $\yt^i$.
The feature at each node can then be updated to $[\z^i, \mathbf{p}^i_{t}$, $\mathbf{m}^i_{t}, \mathbf{s}^i, \mathbf{b}^i]$ before moving on to predict step $t+1$ in the exact same fashion.
Note the autoregressive nature of the decoder allows us to roll out further into the future than just the $6s$ training horizon if desired.

\paragraph{Training}
Training uses a modified CVAE objective:
\begin{align}
    \Lcvae &= \Lrecon + \w{KL} \LKL + \w{coll}\Lcoll \\
    \Lrecon &= \sum_{i=1}^{N} || \Y_\text{post}^i - \Y_\text{gt}^i ||^2 \label{eqn:suppreconloss}\\
    \LKL &= D_\text{KL} (q_\phi(Z | \Ygt, X, \map) || p_\theta(Z | X, \map)) \\
    \Lcoll &= \Lveh + \Lenv.
\end{align}
In practice, $\Lrecon$ only supervises the position and heading, \ie $\Y_\text{post}^i$ and $\Y_\text{gt}^i$ in \cref{eqn:suppreconloss} contain only $[x_t, y_t, \theta_t^x, \theta_t^y]$. The reconstruction loss is applied on one sample from the posterior distribution, while collision losses use a sample from the prior.

$\Lveh$ is introduced in TrafficSim~\cite{suo2021trafficsim}.
It uses a differentiable approximation of vehicle-vehicle collision detection, which represents all $N$ vehicles by discs.
We estimate each agent vehicle $i$ by 5 discs with radius $r_i$, as shown in \cref{fig:collloss}(a), and compute the loss by summing over all pairs of agents $(i,j)$ over time as:
\begin{align}
    \Lveh &= \dfrac{1}{N^2} \sum_{(i,j), i \neq j} \sum_{t=1}^{T} \mathcal{L}_\text{pair}(\yt^i, \yt^j) \label{eqn:vehcoll}\\
    \mathcal{L}_\text{pair}(\yt^i, \yt^j) &= \begin{cases}
                                                1 - \dfrac{d}{r_i + r_j}, & d \leq r_i + r_j \\
                                                0, & \text{otherwise}
                                              \end{cases}
\end{align}
where $d$ is the minimum distance over all pairs of discs representing agents $i$ and $j$.

$\Lenv$ uses a similar idea to penalize collisions with the non-drivable area.
This penalty is only applied to the annotated \emph{ego} vehicle in the nuScenes~\cite{caesar2020nuscenes} sequences during training, since many non-ego vehicles appear off the annotated map.
At each step of rollout, collisions are detected between the ego vehicle and the non-drivable area (by checking for overlap between the rasterized non-drivable layer in $\map$ and the rasterized vehicle bounding box), and a collision point $\mathbf{c}$ is determined as the mean of all vehicle pixels that overlap with non-drivable area.
The loss is then calculated as:
\begin{align}
    \Lenv &= \dfrac{1}{T} \sum_{t=1}^{T} \mathcal{L}_\text{drivable}(\yt, \map) \label{eqn:envcoll} \\
    \mathcal{L}_\text{drivable}(\yt, \map) &= \begin{cases}
                                                    1 - \dfrac{d}{d_\text{max}}, & \text{if partial collision} \\
                                                    0, & \text{otherwise}
                                                  \end{cases}
\end{align}
where $d$ is the distance between the vehicle position (center of bounding box) and collision point $\mathbf{c}$, and $d_\text{max}$ is half the ego bounding box diagonal as shown in \cref{fig:collloss}(b).
Note the loss is only applied if there is a partial collision, \ie only part of the bounding box overlaps with the non-drivable area -- this is because if the vehicle is completely embedded in the non-drivable area, the loss will not give a useful gradient.

The traffic model is implemented in PyTorch~\cite{paszke2017automatic} and trained using the ADAM optimizer~\cite{kingma2014adam} with learning rate $1e^{-5}$ for 110 epochs.
Losses are weighted with $\w{KL} = 4e^{-3}$ and $\w{coll}$ is split into $\w{agent} = 0.05$ for $\Lveh$ and $\w{env} = 0.1$ for $\Lenv$.
The KL loss weight $\w{KL}$ is linearly annealed over time, starting from 0.0 at the start of training up to the full value at epoch 20. 

\subsection{Initialization Optimization}
\label{sec:suppinitoptim}
As discussed in Sec 3.2 of the main paper, initialization of adversarial optimization is done by first performing inference with the learned approximate posterior $q_\phi(Z_\text{init} | Y_\text{init}, X, \map)$ and then running an \emph{initialization optimization} to get the final $Z_\text{init}$ encapsulating both non-ego agents $Z = \{\z^i\}_{i=1}^{N}$ along with the latent planner $\zplan$.

The initialization optimization objective is nearly the same as Eqn 6 in the main paper.
It tries to match the initial future trajectories of non-ego agents (from the input nuScenes scenario) and the ego vehicle (from planner rollout within the initial scenario) that are contained in $Y_\text{init}$:
\begin{equation}
    \min_{Z_\text{init}} \w{match} || Y - Y_\text{init} ||^2 - \w{prior} \log p_\theta(Z_\text{init} | X, \map)
\end{equation}
where $Y = d_\theta(Z_\text{init}, X, \map)$ is the decoded initial scenario but uses \emph{only} position and heading information (\ie no velocities).
For initialization optimization, we use $\w{match} = 10.0$, $\w{prior} = 0.01$, and run for 175 iterations.

\subsection{Adversarial Optimization}
\label{sec:suppadvoptim}
The adversarial optimization is introduced in Sec 3.2 of the main paper. 
Here we provide details about each objective.

\paragraph{Match Planner}
In practice, the objective in Eqn 6 of the main paper only uses the position and heading information contained in $\Ego, \Egoapprox$ (\ie no velocities).
For all experiments, $\alpha = 1e^{-5}$.

\paragraph{Adversarial Loss}
The $\delta$ coefficients in the adversarial objective (Eqn 8 of the main paper) can be explicitly manipulated to discourage certain types of scenarios. 
For all experiments, we dynamically set $\delta_t^i = 0$ if agent $i$ is ``behind'' the planner at time $t$ (\ie we ``mask out'' these agents). 
``Behind'' is determined based on the current heading of the planner: if an agent is outside of the planner's $180 ^{\circ}$ field of view, it is considered behind.
This discourages degenerate scenarios with malicious collisions from behind.

We weight the adversarial loss in Eqn 8 of the main paper by $\w{adv} = 2.0$.

\paragraph{Prior Loss}
In practice, we do not use the $\gamma^i$ coefficients directly to weight the prior loss for each agent.
Instead, we use them to compute a dynamic weight for each agent by interpolating between minimum and maximum hyperparameters $\w{prior}^\text{min}, \w{prior}^\text{max}$.
Eqn 10 from the main paper becomes
\begin{align}
    \Lprior &= - \dfrac{1}{N} \sum_{i=1}^N \log p_\theta(\z^i | X, \map) \cdot \w{prior} (\gamma^i)\\
    \w{prior}(\gamma^i) &= \gamma^i \w{prior}^\text{max} + (1 - \gamma^i) \w{prior}^\text{min}
\end{align}
where we use $\w{prior}^\text{min} = 5e^{-3}$ and $\w{prior}^\text{max} = 1$.
This gives more fine-grained control over the prior loss weight rather than leaving it to $\gamma_i \in [0, 1]$.
In particular, if an agent is a likely adversary (\ie close to colliding with the planner), its weight will be near $\w{prior}^\text{min}$, whereas agents far away will be close to $\w{prior}^\text{max}$.

\paragraph{Initialization Loss}
Similar to the prior loss, the initialization loss (Eqn 12 in the main paper) actually uses $\gamma^i$ to interpolate between max and min hyperparameters and is written:
\begin{align}
    \Linit &= \dfrac{1}{N} \sum_{i=1}^N || \z^i - \z_\text{init}^i ||^2 \cdot \w{init}(\gamma^i) \\
    \w{init}(\gamma^i) &= \gamma^i \w{init}^\text{max} + (1 - \gamma^i) \w{init}^\text{min}
\end{align}
where we use $\w{prior}^\text{min} = 0.05$ and $\w{prior}^\text{max} = 0.5$.

\paragraph{Collision Losses}
The collision term for adversarial optimization is similar to that used to train the traffic model:
\[
\Lcoll = \w{agent}\Lveh + \w{env}\Lenv + \w{plan}\mathcal{L}_\text{plan}.
\]
$\Lveh$ is the same as defined in \cref{eqn:vehcoll} and is applied to all non-ego vehicles to avoid colliding with each other.
$\Lenv$ is the same as defined in \cref{eqn:envcoll} and is applied to all non-ego vehicles (instead of the ego vehicle as in CVAE training) to encourage staying on the drivable area.
$\mathcal{L}_\text{plan}$ is similar to $\Lveh$, but instead of discouraging collisions between non-ego agents, it discourages collisions between the planner and non-ego agents with a large $\gamma^i$ (\ie unlikely adversaries). This term only affects agents that are close to the planner but have large $\gamma^i$ because they have been ``masked out'' as described previously. 
Intuitively, we don't want these agents to ``accidentally'' collide with the planner from behind.

All collision losses are computed on trajectories that are upsampled by $\times 3$ to avoid missing collisions at the low nuScenes rate of 2 Hz. We use $\w{agent} = \w{env} = \w{plan} = 20$.

\paragraph{Optimization}
The adversarial optimization runtime reported in the main paper uses a machine with an NVIDIA Titan RTX GPU and 12x Intel i7-7800X@3.50GHz CPUs.
We optimize for 200 iterations using the ADAM~\cite{kingma2014adam} optimizer (we found L-BFGS is overly prone to local minima for this problem) with learning rate $0.05$.

\subsection{Solution Optimization}
\label{sec:suppsoloptim}
The solution optimization is introduced in Sec 4.1 of the main paper. 
It attempts to find a trajectory for the planner that avoids the collision in the adversarial scenario. 
Like the adversarial optimization, it optimizes $Z$ and $\zplan$ in the latent space of the traffic model and uses very similar objectives:

\paragraph{1. Match Adversarial Scenario}
All non-ego agents should maintain the same trajectories outputted from adversarial optimization.
Let $Y_\text{adv}$ be the set of non-ego trajectories from the collision scenario and $Y$ be the current scenario during solution optimization, then the objective is:
\begin{equation}
    \min_Z || Y_\text{adv} - Y ||^2 - \alpha \log p_\theta(Z | X, \map)
\end{equation}
with $\alpha = 1e^{-4}$.
Note the reason we need to actually optimize $Z$, instead of simply fixing it, is because the non-ego agent trajectories may change as $\zplan$ is optimized due to message passing in the traffic model decoder.

\paragraph{2. Avoid Collisions}
The goal for the planner is to avoid collisions with other agents and the environment while driving plausibly:
\begin{equation}
    \min_{\zplan} \Lprior + \Lcoll.
\end{equation}
The prior loss is
\begin{align}
    \Lprior = - \w{prior}\log p_\theta(\zplan | X, \map)
\end{align}
with $\w{prior} = 5e^{-3}$.
The collision loss is
\[
\Lcoll = \w{agent}\Lveh + \w{env}\Lenv
\]
where $\Lveh$ is the same as defined in \cref{eqn:vehcoll} but only discourages collisions between the planner and all non-ego agents.
$\Lenv$ is the same as defined in \cref{eqn:envcoll} and is applied to the planner only.
We use $\w{agent} = \w{env} = 10$.
When computing collision losses, the planner is rolled out $8s$ into the future (instead of the $6s$ length of the scenario future) to ensure that it does not end up in an irrecoverable state at the end of the scenario.
The planner trajectory is upsampled $\times 3$ before collision checking.

\paragraph{Optimization}
Solution optimization uses the ADAM~\cite{kingma2014adam} optimizer for 200 iterations with a learning rate of $0.05$.

\subsection{Pre-Filtering Potential Scenarios}
\label{sec:suppfeasibility}
As discussed in Sec 5.1 of the main paper, before performing adversarial optimization, initial $8s$ scenarios from nuScenes are filtered to remove those that will be difficult or impossible to cause a collision.
For each potential initialization, 20 futures are sampled from the traffic model conditioned on the past trajectories.
A scenario is considered \emph{feasible} if any of the sampled futures meets the following heuristic conditions, which are designed to find an agent that could be made to collide with the planner:
\begin{itemize}[leftmargin=*] \itemsep0em 
    \item There is a non-ego agent that passes within $10m$ of the planner at some timestep $t$.
    \item That agent is not behind the planner (where ``behind'' is determined in the same way as in \cref{sec:suppadvoptim}) at $t$.
    \item That agent is not separated from the planner by non-drivable map area at $t$. 
    \item The ego vehicle must move $>1$ $m/s$ at some point, avoiding situations where the planner is a ``sitting duck'' with no hope of avoiding a collision.
\end{itemize}
If no samples meet these conditions, the initial scenario is not used for scenario generation.
After doing this feasibility check, we end up with a candidate non-ego agent that could reasonably collide with planner at time $t$.
Note, \name does not use this information in any way -- the adversarial optimization can and does cause collisions with a different agent than the initial candidate.
However, this information is useful for the \emph{Bicycle} baseline which requires the adversary to be chosen before optimization.

\subsection{Bicycle Baseline Scenario Generation}
\label{sec:suppbike}
The \emph{Bicycle} baseline is introduced in Sec 5.1 of the main paper. %
This approach does not use the learned traffic model to parameterize scenarios, instead explicitly optimizing the acceleration profile of an adversary.
As it uses no strong priors on holistic traffic motion, only a single pre-chosen ``attacker'' is optimized (similar to prior work~\cite{wang2021advsim}).
For this, we use the candidate adversary determined by the feasibility check in \cref{sec:suppfeasibility} (in practice, using this feasibility check for \emph{Bicycle} would not be possible since it leverages samples from the traffic model, however we use it here to ensure a fair comparison to \name).

Optimization is performed over the future acceleration profile of the adversary $\A = [\accel_1, \accel_2, \dots, \accel_T]$ with $\accel_t = [\Dot{v}_t, \Ddot{\theta}_t]$.
Let $\mathbf{b}$ be the length and width of the adversary vehicle, then to get the adversary trajectory $\Y = [\y_1, \y_2, \dots, \y_T]$ when needed, the kinematic bicycle model is recursively applied $\yt = \mathcal{K}(\y_{t-1}, \accel_t, \mathbf{b})$ where $\y_0$ is the state at the last timestep of the fixed \emph{past}. 

\paragraph{Initialization Optimization}
Before optimizing to cause a collision, the acceleration profile of the adversary is fit to its corresponding trajectory in the initial nuScenes scenario $\Y_\text{init}$ with
\begin{equation}
    \min_\A || \Y - \Y_\text{init} ||^2.
\end{equation}
This loss is applied only to the position and heading part of the trajectories (\ie not velocities or accelerations).

\paragraph{Adversarial Optimization}
Next, the main adversarial optimization is performed, which encourages the adversary to collide with the planner using
\begin{equation}
    \min_\A \w{adv}\Ladv + \w{accel}\Laccel + \Lcoll.
\end{equation}
\noindent The adversarial term encourages a collision similar to \name by minimizing positional distance: 
\begin{align}
    \Ladv &= \sum_{t=1}^T || \yt - \ego_t || \cdot \delta_t \\
    \delta_t &= \dfrac{\exp(-|| \yt - \ego_t ||)}{\sum_t \exp(-|| \yt - \ego_t ||)} \label{eqn:bikesoftmin}.
\end{align}
The \emph{softmin} in equation \cref{eqn:bikesoftmin} is only choosing a candidate timestep to cause a collision (as opposed to choosing both a candidate agent \emph{and} timestep as in \name) since the colliding agent is fixed ahead of time.
Also note that $\ego_t$ is the actual planner trajectory, not a differentiable approximation from the traffic model as used in \name.
This means that during optimization, it is necessary to compute gradients through the true planner if the setting is closed-loop (\eg the \emph{Rule-based} planner).
As discussed in Sec 5.1 of the main paper, we implemented an explicit gradient estimation using finite differences to be able to backpropagate through the planner.
However, this requires substantially more queries to the planner and is too slow to practically generate many scenarios.
Finite differences, while easy to implement, is inefficient compared to black-box approaches explored in prior work~\cite{wang2021advsim} -- however, it was out of our current scope to explore these options as well since \name does not require them.

The acceleration term regularizes the optimized profile to prefer small accelerations:
\begin{align}
    \Laccel &= \dfrac{1}{T} \sum_{t=1}^T || \accel_t ||^2.
\end{align}

\noindent The collision loss is
\[
\Lcoll = \w{agent}\Lveh + \w{env}\Lenv
\]
where $\Lveh$ is the same as defined in \cref{eqn:vehcoll} but only discourages collisions between the adversary and other (fixed) non-ego agents.
$\Lenv$ is the same as defined in \cref{eqn:envcoll} and is applied to the adversary only.

\paragraph{Optimization Details}
We use $\w{adv} = 1$, $\w{accel} = 1$, and $\w{agent} = \w{env} = 20$.
Initialization optimization uses L-BFGS (which was superior to ADAM for the simple objective) for 50 iterations.
Adversarial optimization uses ADAM for 300 iterations.

To fairly compare to \name and evaluate certain metrics, %
after \emph{Bicycle} adversarial optimization we fit the output scenario within our learned traffic model using the procedure described in \cref{sec:suppinitoptim}.
We can then compute the likelihood of the adversary's $\z$ under the learned prior, and use the exact same solution optimization described in \cref{sec:suppsoloptim}.

\subsection{Rule-based Planner}
\label{sec:suppplanner}
Our rule-based planner is introduced in Sec 4.2 of the main paper and has the following structure:
\begin{enumerate}[leftmargin=*] \itemsep0em 
    \item Extract from the lane graph a finite set of splines that each vehicle might follow.
    \item Generate predictions for the future motion of non-ego vehicles along each of the splines from (1).
    \item Generate candidate trajectories for the ego vehicle and use the predictions from (2) to estimate the ``probability of collision'' $p_{\text{col}}(\tau)$ for each candidate $\tau$.
    \item Among trajectories that are unlikely to collide $\{ \tau \mid p_{\text{col}}(\tau) < p_{\text{max}}\}$, choose the trajectory that covers the most distance. If no trajectories are unlikely to collide, choose the trajectory that is least likely to collide.
    \item Repeat every $\Delta t$ seconds.
\end{enumerate}

Note the ``intent'' of the planner is deterministic, \ie it will always follow the same lane graph path (\eg choosing whether to turn left or right) when rollout starts from the same initialization.
As discussed in the main paper, nuScenes lane graphs do not contain information to switch lanes and since the rule-based planner strictly follows the lane graph, it cannot change lanes.
Planner behavior is affected by hyperparameters such as how $p_{\text{col}}(\tau)$ is computed, $p_{\text{max}}$, and the maximum speed and forward acceleration.

\section{Experimental Details} \label{sec:supp-exptdetails}
In this section, we provide details for the experiments performed in Sec 5 of the main paper.

\subsection{Data and Metrics}
\paragraph{Dataset}
The nuScenes~\cite{caesar2020nuscenes} dataset is used for all experiments; we use the scene splits and settings from the nuScenes prediction challenge.
All vehicle trajectories in the dataset are pre-processed to remove frames where the vehicle bounding box has a $>30\%$ overlap with either the non-drivable area or a car park area.
This avoids scenarios with many auxiliary agents that are off of the annotated map or not moving.
Trajectories are additionally annotated with velocity and yaw rate using finite differences on the provided positions/headings.
Finally, we flip the maps and trajectories for ``Singapore'' data about the $x$ axis so that vehicles across the whole dataset (both in Boston and Singapore) are consistently driving on the right-hand side of the road.

For training the learned traffic model, we use scenes from the training split of the nuScenes prediction challenge.
Note we use all available trajectory data in the training scenes for \emph{cars} and \emph{trucks}, including the ego trajectories and all agents not removed in our own pre-processing.

Scenario generation in all experiments is initialized from a set of 1200 $8s$ nuScenes scenarios (before pre-filtering in \cref{sec:suppfeasibility}) extracted from the train and val splits.
Since the original nuScenes data sequences are $20s$ long, some of these extracted $8s$ scenarios partially overlap.

\paragraph{Metrics}
Acceleration and collision velocity metrics are computed using finite differences.
The accelerations reported in Tab 1 and Tab 3 of the main paper are \emph{forward accelerations} (calculated using the change in speed) while those in Tab 2 are the full acceleration (encompassing both forward and lateral).
This is because Tab 1 and 3 focus on trajectories from the \emph{Rule-based} planner, which follows the lane graph, cannot change lanes, and has a deterministic route as discussed in \cref{sec:suppplanner}.
As a result, generated scenarios cannot make the planner swerve suddenly (which would require leaving the lane graph, changing lanes, or changing route), they can only cause a harsher slow down or speed up, which is captured by measuring forward acceleration.

For all experiments, acceleration (\emph{Accel}), environment collision rate (\emph{Env Coll}), and nearest-neighbor distance (\emph{NN Dist}) are only reported up to the time of collision, since any continuing motion is merely the result of not physically simulating the collision.
For the environment collision rate reported in Tab 2, a vehicle is considered in collision if there is more than 5\% overlap between its bounding box and the non-drivable area.

\subsection{Planner-Specific Scenario Generation}
\label{sec:suppadvgenexpt}
This experiment is presented in Sec 5.1 of the main paper.
During scenario generation, the \emph{Rule-based} planner uses default manually-set hyperparameters (\ie no large-scale tuning was done beforehand, as described in Sec 5.3 of the main paper).
These default hyperparameters were set by observing rollouts on a small subset of nuScenes.

In Tab 1, collision rate is reported over all scenarios for which adversarial optimization was performed (\ie the roughly 500 pre-filtered scenarios).
Solution rate is with respect to all scenarios where a collision was successfully caused, while planner trajectory and match planner metrics are computed over all useful scenarios (those where both a collision and solution were found).

\subsection{Baseline Comparison}
This experiment is presented in Sec 5.1 of the main paper, and compares \name to the baseline scenario generation detailed in \cref{sec:suppbike}.
Metrics reported in Tab 2 are for a set of 139 scenarios where both methods were able to cause a collision from the same nuScenes initialization.
Please see \cref{sec:suppbike} for details on how solution rate and NLL are measured for \emph{Bicycle}.

\subsection{Scenario Analysis}
This experiment is presented in Sec 5.2 of the main paper.
The results shown in Fig 6 are collision labels assigned to the scenarios generated in Sec 5.1.
Note that $k$-means clustering was not performed directly on the scenarios generated in Sec 5.1, instead clustering was done beforehand with a large set of over 400 scenarios generated from various subsets of nuScenes and using many versions of our \emph{Rule-based} planner, giving a wide variety of collisions.
Clusters were assigned semantic labels by visual inspection of the collisions in each cluster.
Then, after generating new scenarios in Sec 5.1, collision types are assigned to each new scenario by simply associating their collision feature with the closest cluster (\ie clustering is \emph{not} done over again, scenarios are assigned to extant clusters).
Videos of representative scenarios assigned to each cluster are shown on the \textbf{supplementary webpage}.

We use $k=10$ for clustering. Finer-grained classification is possible, if necessary, using larger $k$ or additional features like collision velocity, however we found the described approach sufficient to analyze the \emph{Rule-based} planner's performance while being easily interpretable.

\subsection{Improving Rule-based Planner}
This experiment is presented in Sec 5.3 of the main paper.
Before any tuning is performed, the planner starts with the same set of ``default'' hyperparameters described in \cref{sec:suppadvgenexpt}.
When tuning the planner hyperparameters, the optimal set is chosen by lowest collision rate with ties broken by lowest acceleration.
The planner is first tuned on 800 $8s$ nuScenes scenarios (from the train/val splits), which gives initial optimal hyperparameters for ``regular'' driving scenarios.
Adversarial optimization is performed on the planner with both the \emph{default} and \emph{regular-tuned} hyperparameters to create a set of challenging scenarios to guide further improvements.
When tuning on challenging scenarios, ``Behind'' collisions are removed, which we found unrealistic as discussed in Sec 5.2 of the main paper.

The hyperparameters for the \emph{Rule-based} planner are described in \cref{sec:suppplanner}.
Tuning searches over $p_{\text{max}}$ in the range of $[0.05, 0.2]$, max speeds in the range $[12.5, 20.0]$ $m/s$, max accelerations in the range $[3.0, 4.5]$ $m/s^2$, and parameters related to computing $p_{\text{col}}(\tau)$. In total, tuning sweeps over 432 hyperparameter combinations.

Results in Tab 3 of the main paper are on scenarios from the held-out nuScenes test set.
Metrics over collision scenarios (\emph{Coll}) are reported for scenarios where some hyperparameter setting succeeded in avoiding a collision, \ie scenarios where all hyperparameter settings collide are considered impossible and discarded.
The best possible collision rate on regular scenarios is only $3.2\%$ (achieved by choosing a \emph{different} set of parameters for every scenario), making the $4.6\%$ of the regular-tuned planner version close to optimal.
The reason this collision rate cannot be 0 is the use of log replay (\ie rolling out the planner in pre-recorded scenarios), which results in some unavoidable collisions: (i) pre-recorded traffic is not reactive to the planner and (ii) the ego vehicle is sometimes initialized off the lane graph which it cannot robustly handle.

\paragraph{Learned Mode Classifier}
The multi-mode version of the planner uses a binary classifier that decides whether the ego vehicle is currently in a ``regular'' or ``accident-prone'' situation.
In the \emph{learned} version, this classifier is a neural network that has a very similar architecture to the learned traffic model described in \cref{sec:supptrafficarch}.
In particular, it takes in the past $2s$ trajectories for all agents in a scene, along with local map crops around each, and processes them in the same way as the traffic model.
These features are then placed into a scene graph and message passing is performed in the same way as done for the prior.
The output feature (size 64) at the ego node is given to 2-layer MLP that makes a binary classification for the scene.
This network is trained on regular nuScenes scenarios from the training split in addition to a diverse set of over $1000$ collision scenarios generated from train/val scenes using variations of both the \emph{Replay} and \emph{Rule-based} planners.
Training uses a typical binary cross entropy loss that is weighted to account for the data imbalance between regular and collision scenarios.

\section{Supplemental Experiments} \label{sec:supp-results}
In this section, we provide additional results and experiments omitted from the main paper for brevity.

\begin{table}[t]
\begin{center}
\scalebox{1.0}{
\begin{tabular}{l|cc}
\toprule
\multicolumn{1}{c}{\textbf{Model}} & \multicolumn{1}{c}{\textbf{ADE} ($m$) $\downarrow$} & \multicolumn{1}{c}{\textbf{FDE} ($m$) $\downarrow$} \\
\midrule
LDS-AF~\cite{ma2021likelihood} & 1.66 & 3.58 \\
DLow-AF~\cite{yuan2020dlow} & 1.78 & 3.77  \\
Trajectron++~\cite{salzmann2020trajectron++} & 1.51 & - \\
AgentFormer~\cite{yuan2021agent} & \textbf{1.45} & \textbf{2.86} \\
\midrule
Ours, Full & 1.75 & 3.57 \\
Ours, No Bicycle & 1.60 & 3.17 \\
\bottomrule
\end{tabular}}
\end{center}
\vspace{-3mm}
\caption{Learned traffic model future prediction accuracy on all nuScenes prediction categories compared to current state of art. ADE/FDE is reported using 10 samples.}
\label{tab:trafficbaselines}
\end{table}
\begin{table}[t]
\setlength{\tabcolsep}{3pt}
\begin{center}
\scalebox{0.85}{
\begin{tabular}{l|cccc}
\toprule
\multicolumn{1}{c}{\textbf{Model}} & \multicolumn{1}{c}{\textbf{ADE} ($m$)} & \multicolumn{1}{c}{\textbf{FDE} ($m$)} & \multicolumn{1}{c}{\textbf{Env Coll} (\%)} & \multicolumn{1}{c}{\textbf{Veh Coll} (\%)} \\
\midrule
Full  & 1.74 & 3.54 & 10.6 & 5.6 \\
No Bicycle  & \textbf{1.72} & \textbf{3.45} & \textbf{7.2} & \textbf{3.7} \\
No $\Lenv$  & 1.91 & 3.92 & 13.2 & 5.0 \\
No Autoregress & 3.68 & 8.00 & 16.2 & 5.4 \\
\bottomrule
\end{tabular}}
\end{center}
\vspace{-3mm}
\caption{Traffic model ablation study on \emph{cars} and \emph{truck} nuScenes categories only (same as used for scenario generation). Though not using the bicycle model gives better performance, it gives less realistic single-agent vehicle dynamics which is very undesirable for scenario generation.}
\vspace{-5mm}
\label{tab:trafficablation}
\end{table}

\subsection{Traffic Motion Model}
We first evaluate the learned traffic model's ability to accurately predict future motion in a scene.

\paragraph{Data}
We evaluate on the test split of the nuScenes~\cite{caesar2020nuscenes} prediction challenge using $2s$ (4 steps) of past motion to predict $6s$ (12 steps) of future.
This data contains vehicles from the \emph{bus}, \textit{car}, \textit{truck}, \textit{construction}, and \textit{emergency} categories.

\paragraph{Metrics}
Evaluation is done with standard future prediction metrics including minimum average displacement error (\textbf{ADE}) and minimum final displacement error (\textbf{FDE}), which are measured over $K$ samples from the traffic model.
For a single agent being evaluated, these metrics are
\begin{align}
    \text{ADE} &= \min_{k} \dfrac{1}{T} \sum_{t=1}^T || \hat{\y}_t^{(k)} - \y_t ||_2 \\
    \text{FDE} &= \min_{k} || \hat{\y}_T^{(k)} - \y_T ||_2 
\end{align}
where $\hat{\y}_t^{(k)}$ is the predicted \emph{position} of the agent in the $k$th sample at time $t$ and $\y_t$ is the ground truth.
In our experiments, we use $K=10$ samples.

For the ablation study, we also measure the \textbf{environment and vehicle collision rates}. 
Environment collision rate is the fraction of predicted future trajectories where more than $5\%$ of the vehicle bounding box overlaps with the non-drivable area. This is measured over all $K$ samples.
The vehicle collision rate is measured over all agents in each scene (rather than only the single one specified at each data point in the prediction challenge test split) and all $K$ samples.
It is the same as used in TrafficSim~\cite{suo2021trafficsim}, which counts the number of agents in collision (\ie have a bounding box overlap more than IoU 0.02 with another agent).

\vspace{-2mm}
\subsubsection{Baseline Comparison}
\vspace{-1mm}
Prediction performance is compared to reported results for recent state-of-the art models AgentFormer~\cite{yuan2021agent}, Trajectron++~\cite{salzmann2020trajectron++}, DLow-AF~\cite{yuan2020dlow}, and LDS-AF~\cite{ma2021likelihood}.
Results are shown in \cref{tab:trafficbaselines}.
Our full model is trained only on $car$ and $truck$ vehicles to be used for scenario generation, so to evaluate on the prediction challenge test split, we modify the category of input vehicles to our model to be one of these (\eg \emph{bus} $\rightarrow$ \emph{truck}).
Our learned traffic model makes accurate predictions and is competitive with current SOTA methods as shown in \cref{tab:trafficbaselines}.
We also train an ablation of our model that does not use the kinematic bicycle model, instead the decoder directly predicts output position and headings.
This version is trained on all categories in the challenge dataset, and makes more accurate predictions according to ADE/FDE.
Note, however, that using the bicycle model is very important for adversarial and solution optimization to ensure output trajectories have reasonable dynamics even when the optimized $Z$ is off-manifold.

\begin{table*}
\begin{center}
\scalebox{0.825}{
\begin{tabular}{ll|cc|cccc|ccc}
\toprule
\multicolumn{4}{c}{} & \multicolumn{4}{c}{\textbf{Plausibility of Adversary Trajectory} $\downarrow$} & \multicolumn{3}{c}{\textbf{Plausibility of Other Trajectories } $\downarrow$} \\
\multicolumn{1}{c}{\textbf{Scenarios}} & \multicolumn{1}{c}{\textbf{Objective}} & \multicolumn{1}{c}{Col \small{(\%)}} & \multicolumn{1}{c}{Sol \small{(\%)}} & Accel \small{($m/s^2$)} & \multicolumn{1}{c}{Env Coll \small{(\%)}} & NN Dist \small{(m)} & \multicolumn{1}{c}{NLL} & Accel \small{($m/s^2$)} & NN Dist \small{(m)} & \multicolumn{1}{c}{NLL} \\
\midrule
All methods & Full & - & \textbf{87.9} & \textbf{1.11} & \textbf{9.4} & 0.90 & 517.0 & \textbf{0.45} & 0.39 & 341.4 \\
collide & No $\Lprior$ & - & 78.8 & 1.26 & \textbf{9.4} & 1.05 & 502.9 & 0.47 &  0.39 & 474.2 \\
& No $\Lcoll$ & - & 81.8 & 1.43 & 15.6 & 1.11 & 487.0 & \textbf{0.45} & 0.39 & 351.5 \\
& No $\Linit$ & - & 78.8 & 1.19 & \textbf{9.4} & 0.83 & 579.2 & 0.51 & \textbf{0.36} & 21.8 \\
& No $\Linit$, $\gamma$ & - & 84.8 & 1.13 & \textbf{9.4} & \textbf{0.76} & \textbf{89.2} & 0.49 & 0.40 & \textbf{-2.9} \\
& No $\Linit$, $\gamma$, $\delta$  & - & 57.6 & 1.38 & 15.6 & 0.99 & 154.6 & 1.04 & 0.65 & 104.3 \\
\midrule
All collision & Full & 27.4 & 86.8 & 1.30 & 13.3 & 0.95 & 555.9 & 0.43 & 0.35  & 365.2 \\
scenarios for & No $\Lprior$ & 27.2 & 83.9 & 1.29 & 19.1 & 0.95 & 571.8 & 0.39 & 0.31 & 486.0 \\
each method & No $\Lcoll$ & 38.0 & 82.1 & 1.42 & 19.6 & 0.91 & 540.8 & 0.38 & 0.27 & 357.0 \\
& No $\Linit$ & 34.6 & 80.1 & 1.40 & 18.3 & 1.02 & 611.6 & 0.41 & 0.29 & 6.8 \\
& No $\Linit$, $\gamma$ & 30.5 & 76.1 & 1.34 & 14.2 & 0.97 & 110.4 & 0.40 & 0.30 & 3.0 \\
& No $\Linit$, $\gamma$, $\delta$  & 45.9 & 58.1 & 2.01 & 26.4 & 1.38 & 230.8 & 1.07 & 0.71 & 127.9 \\
\bottomrule
\end{tabular}}
\end{center}
\vspace{-3mm}
\caption{Ablation study on adversarial optimization objective for scenario generation on the \emph{Rule-based} planner. In the top section, metrics are reported only for scenarios where all methods were able to cause a collision. In the bottom section, metrics for each method are computed over all collision scenarios generated by that method only, \ie are not directly comparable. Therefore, collision rate is also reported for reference and numbers are not bolded.}
\label{tab:suppadvgenablation}
\vspace{-5mm}
\end{table*}

\subsubsection{Ablation Study}
\label{sec:supptrafficablation}
To evaluate key design differences from TrafficSim~\cite{suo2021trafficsim} and ILVM~\cite{casas2020implicit}, which our model is based on, we ablate various components of our traffic model design.
Results are shown in \cref{tab:trafficablation}, where all models are trained and evaluated only on the $car$ and $truck$ categories, since this is what we use in the main paper for scenario generation.
Same as \cref{tab:trafficbaselines}, \emph{No Bicycle} directly predicts the position and heading from the decoder rather than acceleration profiles that go through the kinematic bicycle model; again, this gives slightly improved performance but less realistic per-agent dynamics.
\emph{No $\Lenv$} only uses vehicle collision penalties while training, similar to prior work~\cite{suo2021trafficsim}.
Removing the environment collision penalty results in a higher collision rate and lower predictive accuracy.
\emph{No Autoregress} uses a GNN decoder that predicts the entire future trajectory in one shot rather than as an autoregressive rollout.
This makes the future prediction task more difficult, substantially reducing accuracy.

\subsection{Adversarial Optimization Ablation Study}
Next, we study how various components of the adversarial optimization objective function (introduced in Sec 3.2 of the main paper and detailed in \cref{sec:suppadvoptim}) affect the generated scenarios.
We consider the following variations:
\begin{itemize}[leftmargin=*] \itemsep0em 
    \item No $\Lprior$ -- removes the prior loss which keeps agents likely under the learned prior.
    \item No $\Lcoll$ -- removes both environment and vehicle collision penalties.
    \item No $\Linit$ -- removes the initialization loss which keeps agents near the initial (realistic) scenario.
    \item No $\Linit,\gamma$  -- removes the $\gamma$ weights from $\Lprior$, meaning likely adversaries will need to stay just as likely as all other agents in the scenario.
    \item No $\Linit,\gamma,\delta$  -- additionally removes the $\delta$ weighting scheme from $\Ladv$, meaning all agents will be simultaneously trying to collide with the planner at all timesteps, and it is left entirely up to $\Lprior$ to avoid unrealistic many-vehicle pileups. 
\end{itemize}
Note that when ablating the $\delta$ and $\gamma$ weightings, we also remove $\Linit$ because adversaries should not be expected to stay very close to initialization when needing to collide.

Results are shown in \cref{tab:suppadvgenablation}.
We use the same metrics as for the baseline comparison in Sec 5.1 of the main paper.
In addition to computing metrics for the adversary (colliding agent), results for all other non-planner agents are also reported (except for environment collision since nuScenes contains many agents driving off the annotated drivable area).
These metrics are not perfect, and it can be hard to evaluate which optimization objective produces ``better'' scenarios (or even impossible since desired characteristics are dependent on downstream use), however they do give insight into the kinds of scenarios being produced.

The top section of \cref{tab:suppadvgenablation} computes metrics over scenarios where all methods caused a collision (same protocol as Tab 2 in the main paper).
However, since there are many variations, this is only 33 scenarios in total.
To give a more complete picture of each variation, the bottom section of the table computes metrics over all generated collision scenarios for each method. It also reports the collision rate to contextualize the solution rate and other metrics.
Because metrics are computed over a different set of scenarios for each method in the bottom section, numbers are not directly comparable, however trends often mirror those in the top part.

\begin{figure}[t]
  \centering
  \includegraphics[width=\linewidth]{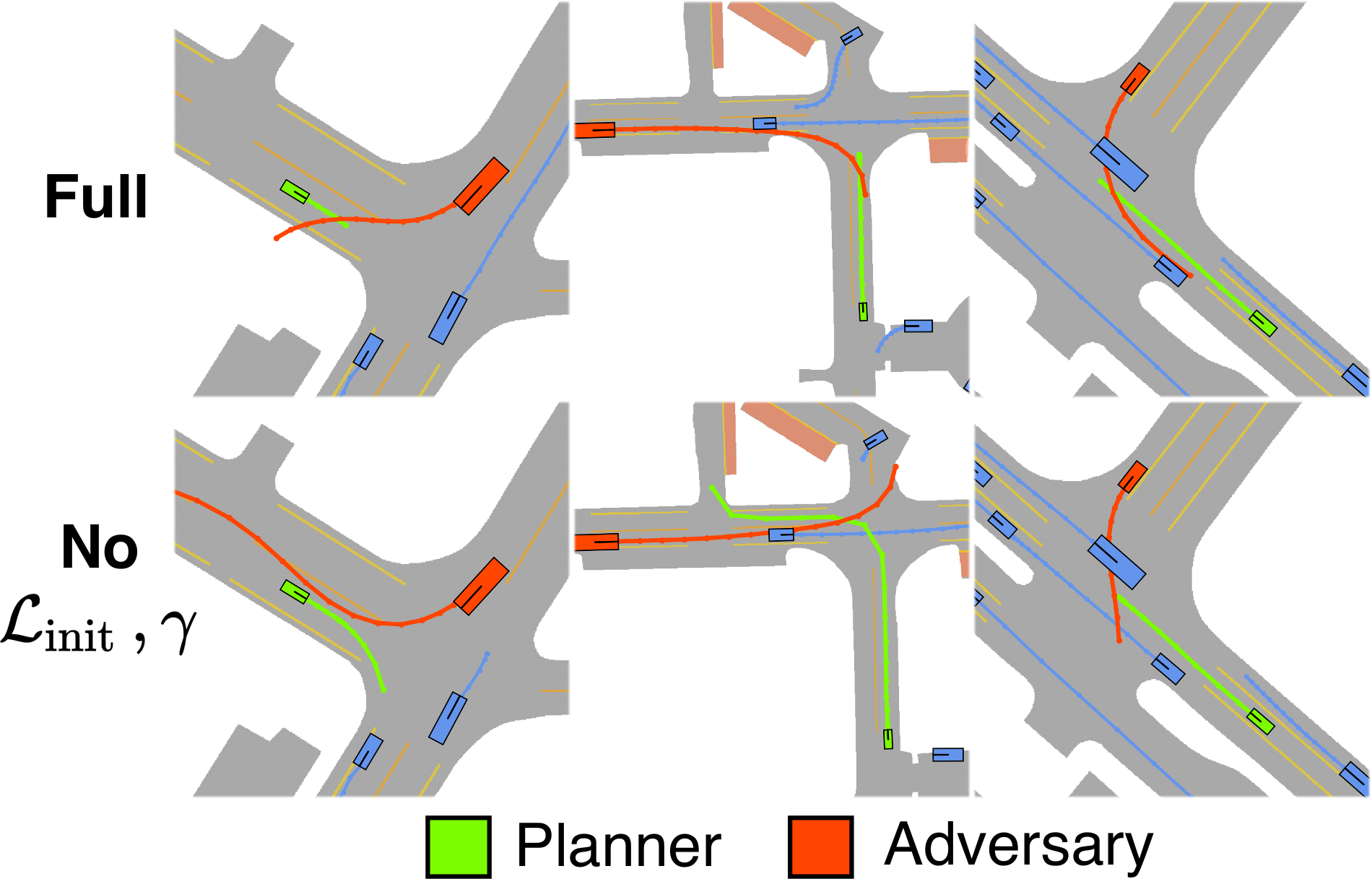}
  \vspace{-4mm}
  \caption{Comparison of generated adversarial scenarios for the \emph{Rule-based} planner using the full objective function and a modified objective with no initialization loss or per-agent weighting in $\Lprior$. Removing per-agent weighting requires adversaries to be more likely under the prior, which can cause collisions that are more aligned with ``usual'' traffic. \name gives the flexibility to modify this objective to generate scenarios best suited for downstream applications.}
  \vspace{-5mm}
  \label{fig:advgenablation}
\end{figure}

In the top of \cref{tab:suppadvgenablation} we see the full objective gives the highest solution rate, \ie it generates \emph{useful} scenarios at the highest frequency.
\emph{No $\Lprior$} tends to cause less likely trajectories for other agents in the scene since they are no longer constrained by the learned prior, and make environment collisions more common for the adversary as seen in the bottom section.
\emph{No $\Lcoll$} similarly causes far more collisions in addition to adversaries with higher accelerations.
Despite this, trajectories maintain reasonable likelihoods since $\Lprior$ is still used and the traffic model does allows for some collisions as seen in \cref{sec:supptrafficablation}.
\emph{No $\Linit$} allows trajectories to stray far from the nuScenes initialization, which produces solvable scenarios at a lower rate and less plausible adversary motion in the bottom part of the table. 
Note that \emph{NLL} for ``others'' is trivially very low since the only remaining regularization is $\Lprior$.
\emph{No $\Linit,\gamma$} forces adversaries to stay more likely under the prior resulting in a low NLL, but lowered solution rate.
The reasonable results produced by this variation indicate the flexibility of our formulation to produce different kinds of scenarios.
By removing $\gamma$, we encourage scenarios where collisions happen within a more ``typical'' setting, rather than as a result of out-of-distribution, adversarial behavior.
This is shown qualitatively in \cref{fig:advgenablation}.
\emph{No $\Linit,\gamma,\delta$} enables many adversaries to attack simultaneously producing many unsolvable scenarios with unrealistic trajectories.

\begin{figure}[t]
  \centering
  \includegraphics[width=0.9\linewidth]{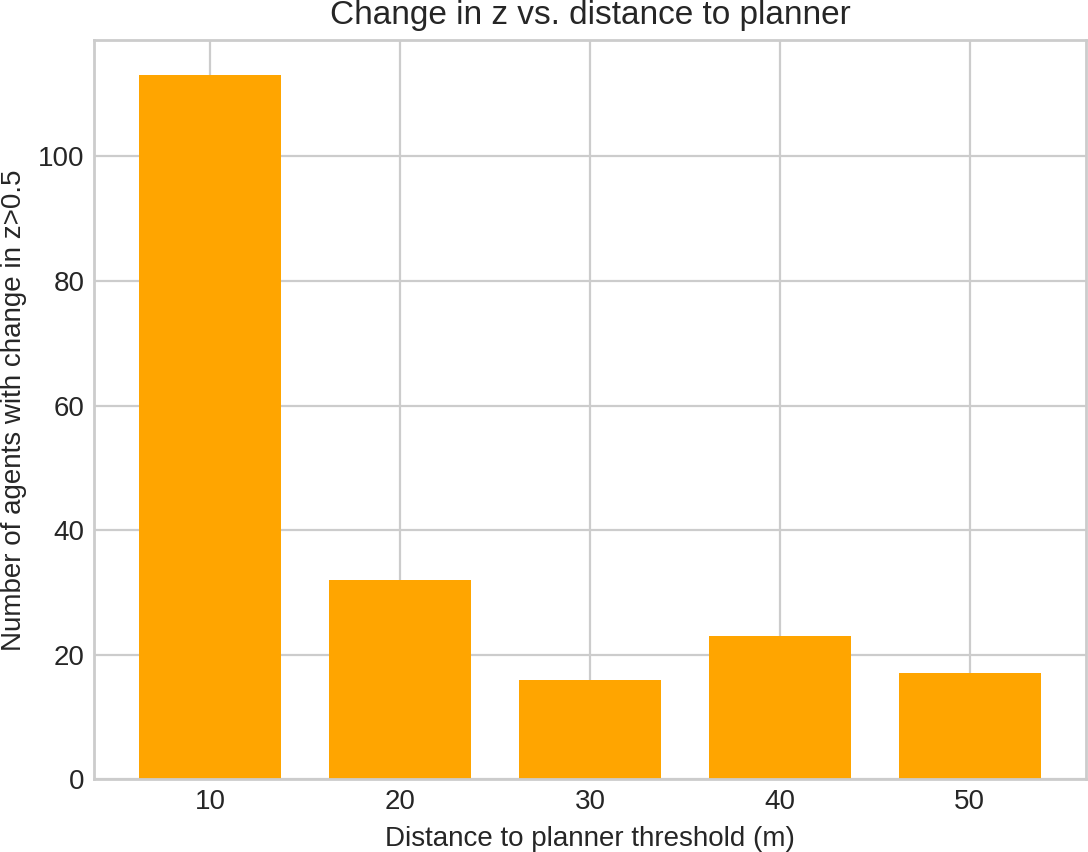}
  \vspace{-1mm}
  \caption{Change in per-agent latent vectors resulting from adversarial optimization as a function of distance to the planner. The bottom axis indicates distance intervals from the planner (\eg 10 indicates 0-10$m$, 20 from 10-20$m$, etc..). For each interval, the number of agents whose latent distance from before to after adversarial optimization is $>~0.5$ is reported (see text for details). We observe that adversarial optimization tends to greatly affect agents that are close to the planner.}
  \vspace{-4mm}
  \label{fig:manyagentresults}
\end{figure}

\subsection{Performance on Many-agent Scenes}
To evaluate the ability of \name to effectively optimize large scenarios, we look specifically at scenes that contain many agents.
The mean number of agents in nuScenes is $11.4$, so it does not contain many massive scenes: only $2/496$ optimized scenarios in Tab 1 of the main paper for the \emph{Rule-based} planner contain $>$~$50$ agents.
For scenarios with $\geq$~$20$ agents, the collision rate is $21.2\%$, which is consistent with $27.4\%$ in Tab 1, indicating performance on large-scale scenes is similar to smaller ones.
This is because even when a scenario has many agents, only a handful near the planner will greatly affect the outcome (due to the optimization objective in Eq 9 of the main paper).
As shown in \cref{fig:manyagentresults}, adversarial optimization makes large changes to the latent vectors of agents within $10m$ of the planner much more frequently than for distant agents.
The change in latent $\z^i$ for each agent is measured as the Euclidean distance between the initialized latent and the final latent (after adversarial optimization), normalized by the maximum difference observed in the scene.

\subsection{Modeling Second-order Effects}
One interesting ability of adversarial optimization is to produce scenarios that cause collisions with the planner using so-called ``second-order effects."
In these scenarios, an adversary may not directly collide with the planner in an obviously malicious way -- instead, it performs some action that causes the planner or other vehicles to react, thereby causing a collision between the planner and some other agent.
For example, we have observed several instances of an adversary in front of the planner suddenly braking, causing the planner to brake and get hit by another agent from behind, rather than the initial adversary in front.
This prompted the regularizer $\mathcal{L}_\text{plan}$ introduced in \cref{sec:suppadvoptim}.
The second-order behavior is possible since \name optimizes \textit{all} agents in a scene rather than choosing specific adversaries ahead of time.
Example videos of these second-order scenarios are shown in the supplementary webpage.

\subsection{Solution Optimization Evaluation}
To confirm the solution optimization is filtering out overly-difficult scenarios, we evaluate using the hyperparameter tuning procedure and generated challenging scenarios introduced in Sec 5.3 of the main paper.
In particular, we take the vanilla \emph{Rule-based} planner and perform two hyperparameter tuning sweeps: one on generated collision scenarios where the solution optimization found a solution, and one on collision scenarios where no solution could be found.
For each sweep, we measure the mean fraction of hyperparameter settings that succeed (\ie don't collide) per scenario in the tuning set.
When tuning on \textit{solution-failed} scenarios, \textbf{only $11.8\%$ of hyperparameter combinations succeed} in avoiding collisions for each scenario on average.
However for tuning on \textit{solution-found} scenarios, $29.8\%$ of hyperparameters succeed for each. 
This indicates that finding a sufficient hyperparameter setting for scenarios where our solution optimization failed is more difficult than for those where a solution could be found.

\subsection{Additional Qualitative Results}
Additional qualitative results of \name on the \emph{Rule-based} planner are shown in \cref{fig:advgenqual}.
Video examples are also included in the supplementary webpage.

\begin{figure}[t]
  \centering
  \includegraphics[width=1.0\linewidth]{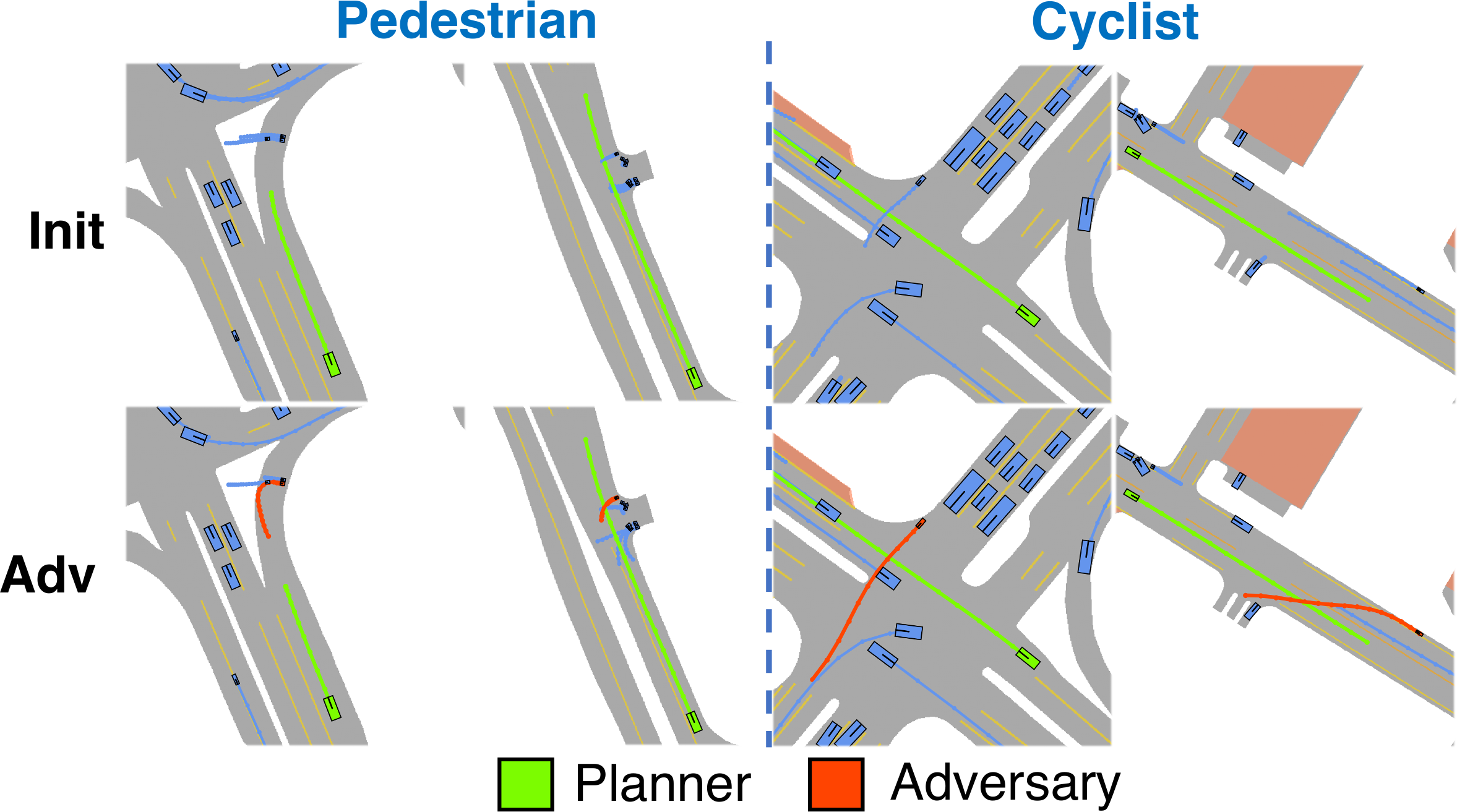}
  \vspace{-3mm}
  \caption{Generated scenarios for the \emph{Replay} planner using a traffic model trained on \emph{all} categories. Top row shows the initial scene and bottom is the output of adversarial optimization. When choosing the adversary ahead of time, \name can cause collisions with both pedestrians (left) and cyclists (right).}
  \vspace{-5mm}
  \label{fig:pedexample}
\end{figure}

\subsection{Pedestrian and Cyclist Adversaries}
Though our main focus in this work is generating scenarios involving vehicles, as a proof-of-concept we train the learned traffic model on all categories in the nuScenes dataset and generate scenarios for the \emph{Replay} planner where the adversary is a pedestrian or cyclist.
For this experiment, a pedestrian/cyclist adversary is specifically chosen before optimization using the procedure in \cref{sec:suppfeasibility}.
Results are shown in \cref{fig:pedexample} and on the webpage.

\begin{figure}[t]
  \centering
  \includegraphics[width=\linewidth]{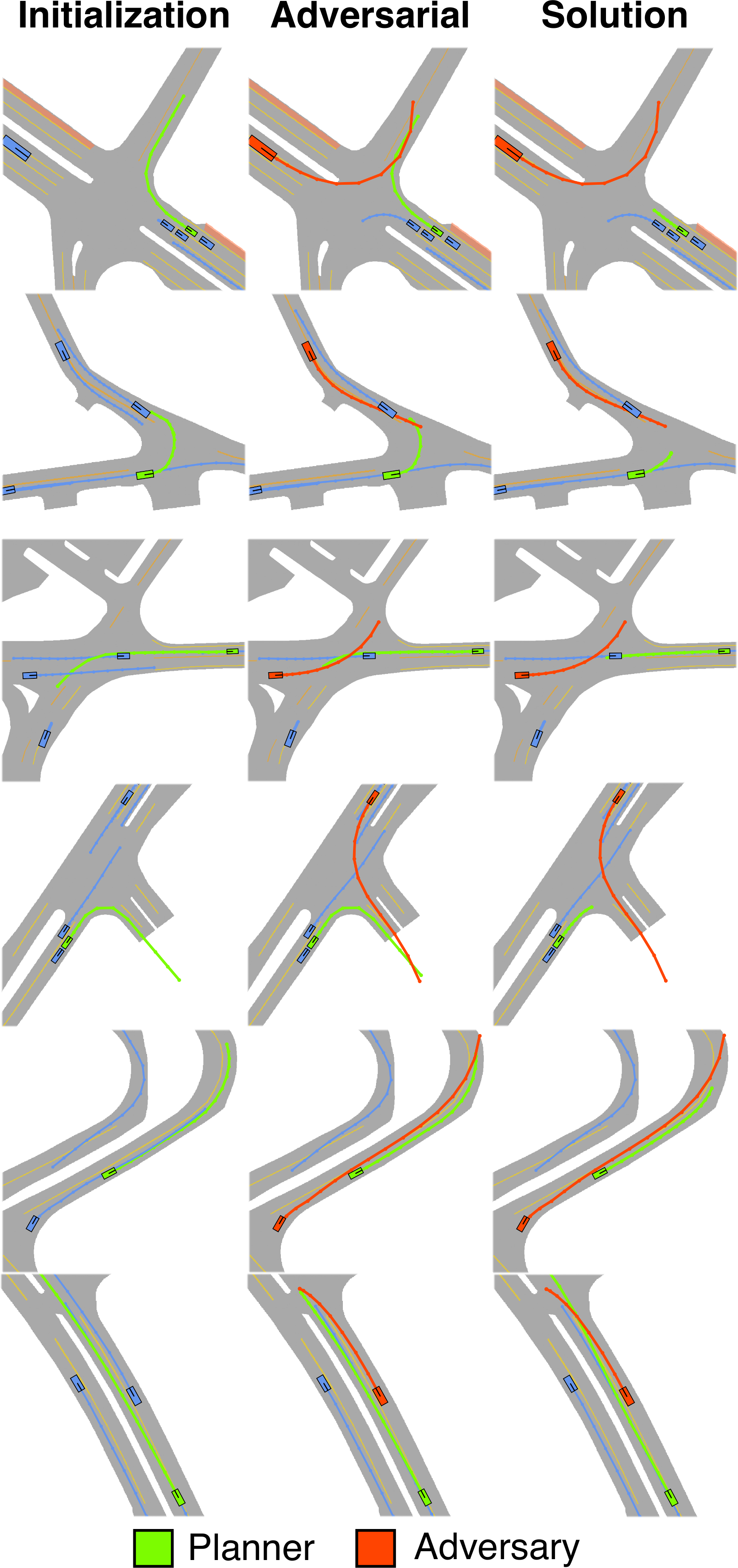}
  \vspace{-4mm}
  \caption{Additional qualitative results of \name for the closed-loop \emph{Rule-based} planner. Adversarial optimization makes large changes to the initial scenario from nuScenes, \eg changing the intent of the adversary or moving stationary vehicles, to cause useful collision scenarios.}
  \vspace{-5mm}
  \label{fig:advgenqual}
\end{figure}

\subsection{Failure Cases and Limitations}
\begin{figure}[t]
  \centering
  \includegraphics[width=0.7\linewidth]{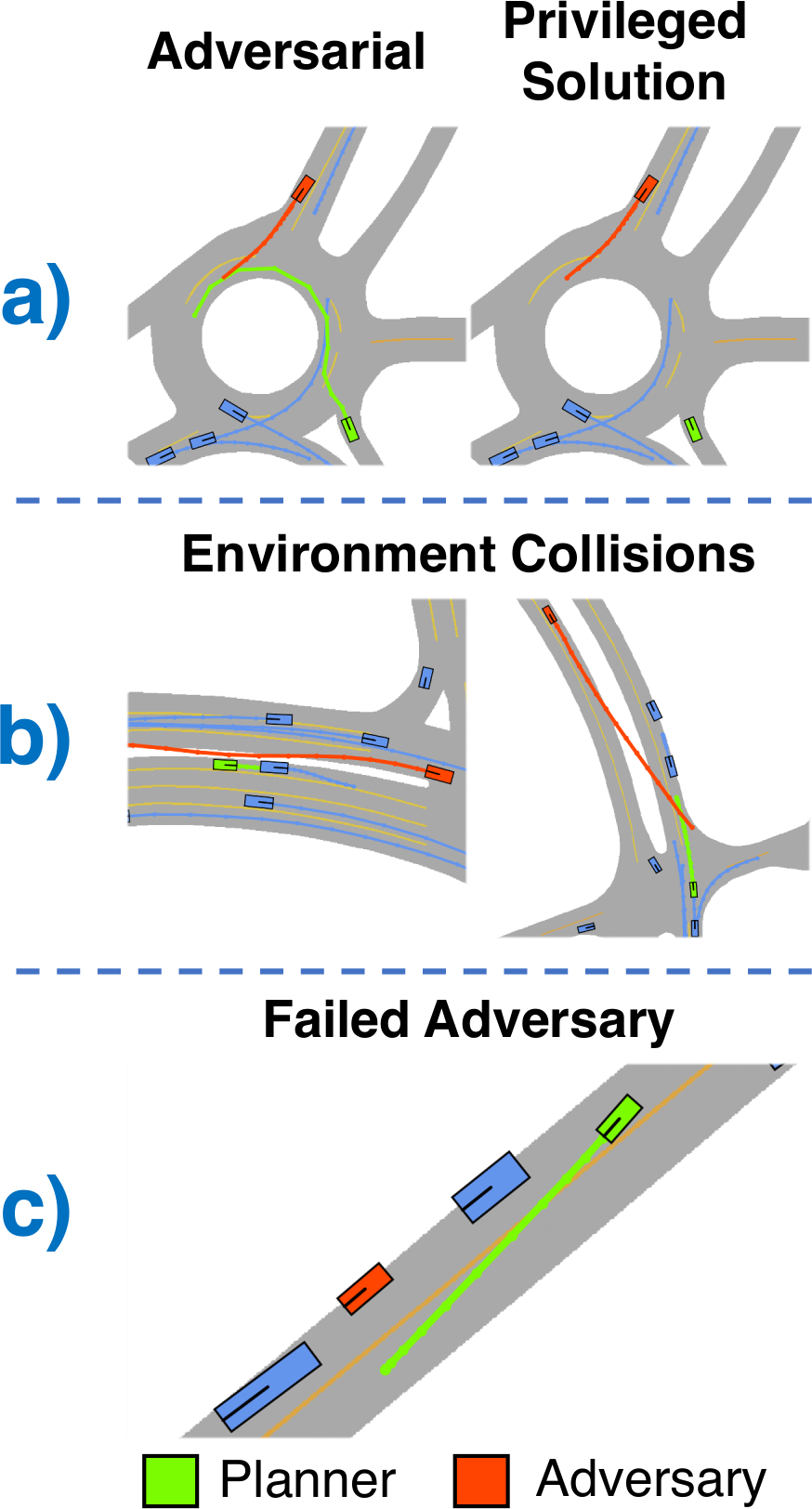}
  \vspace{-1mm}
  \caption{Example failure cases of \name. (a) The solution optimization has access to privileged information, sometimes resulting in unrealistic ``solutions''. (b) Adversaries sometimes drive on non-drivable area to cause a collisions. (c) Attacks that require unlikely motion under the learned prior (\eg a parked car pulling out) can be difficult to produce.}
  \vspace{-5mm}
  \label{fig:failures}
\end{figure}

In addition to the limitations discussed in Sec 6 of the main paper, \cref{fig:failures} shows examples of other \name limitations.
First, our proposed solution optimization is iterative and operates on the full temporal planner trajectory, therefore it has access to future information that sometimes allows performing evasive maneuvers even before an ``attack'' is apparent.
An example is in \cref{fig:failures}(a) where the optimized solution simply does not not pull into the roundabout where the collision occurs.
\cref{fig:failures}(b) shows that the adversary sometimes crosses non-drivable areas in order to collide with the planner.
Though this scenario is technically possible, it is extreme behavior that may not be desired.
However, these situations can be easily detected and discarded, and usually occur only when there is no other feasible adversary near the planner.
Finally, adversarial optimization can have difficulty exhibiting behavior that is very unlikely under the prior even when it is realistic, \eg a parked car pulling out as shown in \cref{fig:failures}(c), since these motions are rare in the traffic model training data.

\end{document}